\documentclass[11pt]{article}

\usepackage[final]{acl}

\usepackage{times}
\usepackage{latexsym}

\usepackage[T1]{fontenc}

\usepackage[utf8]{inputenc}

\usepackage{microtype}

\usepackage{inconsolata}

\usepackage{graphicx}
\usepackage{tabularx}
\usepackage{booktabs}
\usepackage{graphicx}
\usepackage{amssymb}
\usepackage{amsmath}
\usepackage{enumitem}
\usepackage[table]{xcolor}
\usepackage[most]{tcolorbox}
\usepackage[dvipsnames]{xcolor}
\usepackage{xcolor}
\usepackage{multirow}
\usepackage{hyperref}
\usepackage{fontawesome5}

\tcbset{
  mygraybox/.style={
    colback=gray!10,   
    colframe=gray!60,   
    boxrule=0.5pt,
    arc=2pt,
    left=6pt,right=6pt,top=6pt,bottom=6pt
  }
}

\title{Do LLM Self-Explanations Help Users Predict Model Behavior?
Evaluating Counterfactual Simulatability with Pragmatic Perturbations}

\author{
\textbf{Pingjun Hong \textsuperscript{\faLaptopCode\kern1pt\faGraduationCap}},
\textbf{Benjamin Roth \textsuperscript{\faLaptopCode\kern1pt\faBook}}
\\[8pt]
\textsuperscript{\faLaptopCode} Faculty of Computer Science, University of Vienna, Vienna, Austria\\
\textsuperscript{\faGraduationCap} UniVie Doctoral School Computer Science, University of Vienna, Vienna, Austria\\
\textsuperscript{\faBook} Faculty of Philological and Cultural Studies, University of Vienna, Vienna, Austria\\
\tt{
\{\href{mailto:pingjun.hong@univie.ac.at}{\textcolor{black}{pingjun.hong}},
\href{mailto:benjamin.roth@univie.ac.at}{\textcolor{black}{benjamin.roth}}\}@univie.ac.at}
}

\begin{document}
\maketitle
\begin{abstract}
Large Language Models (LLMs) can produce verbalized self-explanations, yet prior studies suggest that such rationales may not reliably reflect the model’s true decision process. We ask whether these explanations nevertheless help users predict model behavior, operationalized as \textit{counterfactual simulatability}. Using StrategyQA, we evaluate how well humans and LLM judges can predict a model’s answers to counterfactual follow-up questions, with and without access to the model’s chain-of-thought or post-hoc explanations. We compare LLM-generated counterfactuals with pragmatics-based perturbations as alternative ways to construct test cases for assessing the potential usefulness of explanations. Our results show that self-explanations consistently improve simulation accuracy for both LLM judges and humans, but the degree and stability of gains depend strongly on the perturbation strategy and judge strength. We also conduct a qualitative analysis of free-text justifications written by human users when predicting the model’s behavior, which provides evidence that access to explanations helps humans form more accurate predictions on the perturbed questions.
\end{abstract}

\section{Introduction}

\begin{figure}[t]
  \includegraphics[width=\columnwidth]{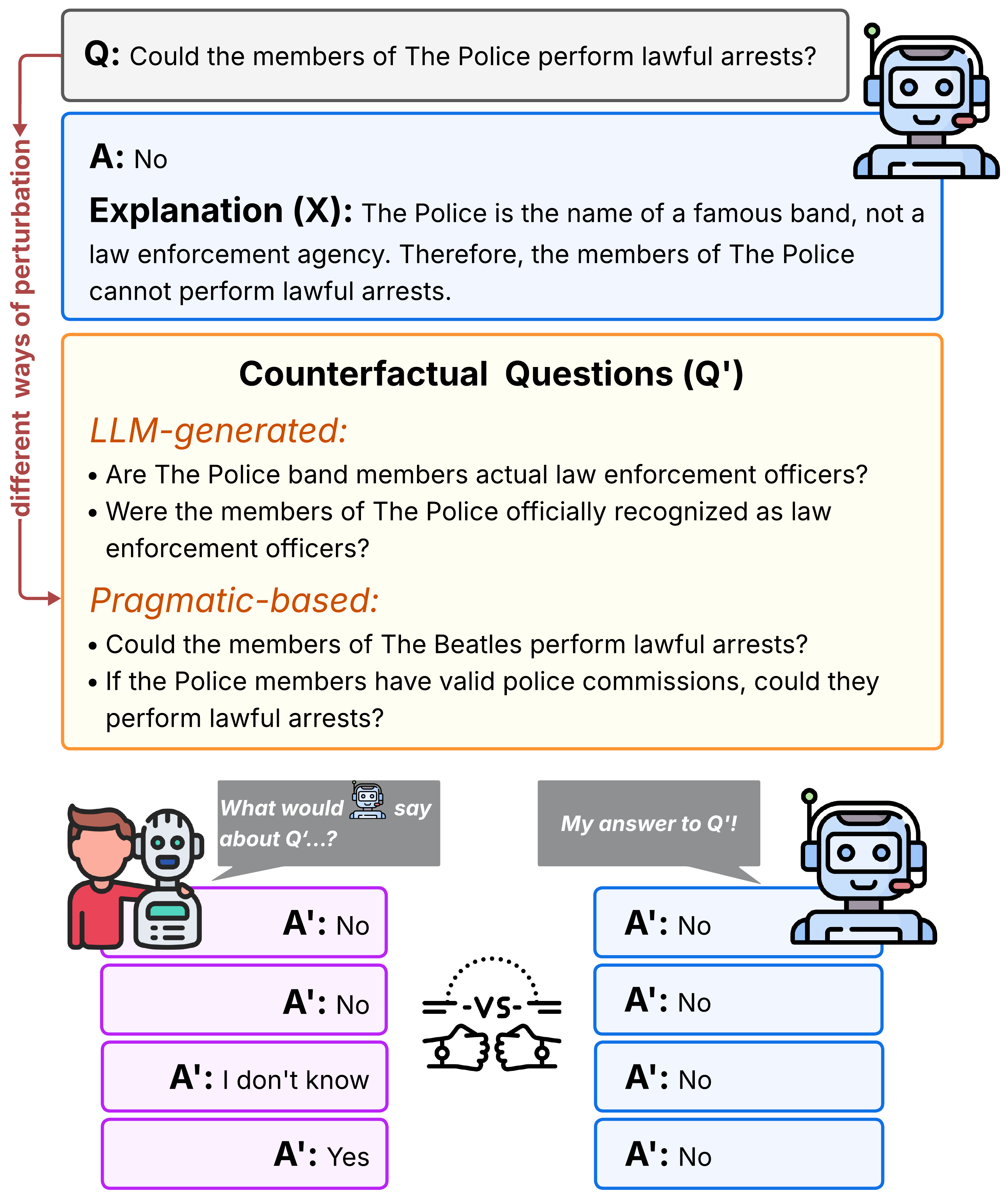}
  \caption{Pipeline for evaluating the usefulness of LLM self-explanations based on different counterfactual question perturbations. For each original $Q$ and model prediction, we measure users’ ability to predict model behavior on different counterfactual question sets $Q'$, with and without access to the model’s self-explanations.}
  \label{fig:intro}
\end{figure}

Large language models (LLMs) can generate \textit{verbalized self-explanations} that accompany a model’s decision. These self-explanations are increasingly used as a window into how the model “reasons”. Meanwhile, a growing body of work has raised concerns about the \textit{faithfulness} of such explanations, showing that these rationales can be decoupled from the underlying decision process, or manipulated without affecting the model’s predictions, suggesting that self-explanations should not be trusted as accounts of model behavior \citep{atanasova-etal-2023-faithfulness,madsen-etal-2024-self}. This leaves open a question for explainable AI: even if LLM verbalized self-explanations are not strictly faithful, can they nonetheless \emph{contribute to model understanding} by helping users form more accurate predictions about how the model will behave? Recent works have proposed to assess explanation quality through \textit{counterfactual simulatability}: an explanation is considered useful if it helps people anticipate how a model’s prediction would change on counterfactual inputs \citep{DoshiVelez2017TowardsAR, chen2024,limpijankit-etal-2025-counterfactual}.

Natural language counterfactuals, defined as minimal \textit{what-if} variations of an input \citep{Miller2017ExplanationIA, ross-etal-2021-explaining, inbook}, are widely used in explainable AI to reveal how models behave under controlled perturbations \citep{chen2024,wang-etal-2024-coxql,wang-etal-2025-fitcf}. In a counterfactual simulatability protocol, users see an original question $Q$ and the model’s prediction $A$, then predict the model’s outputs on  counterfactual variants $Q'$. The usefulness of an explanation is measured by how much it improves the simulation accuracy.

While this line of work offers a promising operationalization of explanation usefulness, much less attention has been paid to the \textit{counterfactuals themselves}. We still lack a systematic understanding of how different ways of perturbing a question affect the conclusions about the usefulness of explanations. In \citet{chen2024}'s setup, for example, the counterfactual questions are generated by LLMs using prompt-based designs. If the evaluation protocol is sensitive to the choice of perturbation, the measured \textit{usefulness} of an explanation may largely be an artifact of how those counterfactuals are constructed.

Figure~\ref{fig:intro} outlines our pipeline. We extend the \textit{counterfactual simulatability} framework to assess how LLM verbalized self-explanations support model understanding. We study \textit{when} and \textit{how} explanations help under different question perturbations. Specifically, we generate counterfactual variants $Q'$ of an original question $Q$ and measure explanation usefulness by how much explanations help users predict the model’s behavior on $Q'$. Perturbations that yield larger gains form a more informative testbed for evaluating self-explanations.

To summarize, our paper addresses the following research questions:
\textbf{(i)} Do LLM verbalized self-explanations contribute to model understanding as measured by counterfactual simulatability? 
\textbf{(ii)} How do different counterfactual perturbation strategies assess the usefulness of LLM self-explanations
and the difference between conditions with and
without explanations?
\textbf{(iii)} Are users biased toward the model's original answer when they reason about $Q'$, and can self-explanations reduce this bias?
\textbf{(iv)} How do self-explanations affect human users’ confidence and accuracy in predicting model behavior, and to what extent do LLM-as-a-judge predictions align with human judgments?

\section{Related Work} \label{sec:related-work}
\paragraph{Evaluating Explanations Simulatability.} 
\textit{Simulatability} measures how well humans can predict a model’s outputs based on its explanations \citep{DoshiVelez2017TowardsAR, ribeiro-etal-2016-trust, chandrasekaran-etal-2018-explanations, hase-bansal-2020-evaluating, chen2024}. \textit{Faithfulness}, as a common metric for explanations, measures whether an explanation is consistent with the model’s decision process \citep{Gilpin2018ExplainingEA, AlvarezMelis2018TowardsRI, Jacovi2020TowardsFI}. Following \citet{chen2024}, simulatability can be seen as a special case of faithfulness, where the output predictor is restricted to humans or LLM judges instead of any arbitrary black box model. In our work, we focus on simulatability as a way to capture the usefulness of explanations \citep{DoshiVelez2017TowardsAR, Hoffman2018MetricsFE}.

Different from \citet{chen2024}, who run the simulatability setting once with explanations, we conduct a forward simulation experiment with two phases, following the experimental setup of the previous work, where they test whether users can better predict a detector’s behavior after seeing explanations \citep{hase-bansal-2020-evaluating,DoshiVelez2017TowardsAR,Schoenegger2024AnEO}. We run experiments with both human users and LLM judges to evaluate how useful the model’s own verbal explanations are for predicting its behavior.

\paragraph{Counterfactual Generation.}
Counterfactual reasoning is the process of thinking about what would have happened if conditions were different, and it is a common tool for judging causality \citep{Kahneman1982TheSH}. Such causal judgments are important for model evaluation, error analysis, and explanation \citep{Miller2017ExplanationIA}. In NLP, most work defines a relationship between an original input $x$ and a changed version $\hat{x}$, then constructs $\hat{x}$ to follow this relationship \citep{wu-etal-2021-polyjuice}. For example, \citet{gardner-etal-2020-evaluating} ask dataset authors to manually edit test examples in small but meaningful ways that usually flip the gold label.

However, manual counterfactuals are costly, so later work turns to model-based generation. \citet{wu-etal-2021-polyjuice} fine-tune GPT-2 \citep{Radford2019LanguageMA} to produce counterfactual examples given a desired edit type. \citet{wang-etal-2025-fitcf} propose FITCF, a framework that filters counterfactuals by checking whether the label flips, then uses them as demonstrations for few-shot prompting, achieving better results than three strong baselines. In \citet{chen2024}, which is closest to our setup, LLMs are prompted to generate counterfactual questions for \textbf{StrategyQA}. In contrast, we compare several counterfactual generation methods and propose a more controlled, pragmatics-based editing approach, and we compare it with counterfactual questions produced by generative models.

\section{Pragmatic Counterfactual Taxonomy} \label{sec:pragcf}

As in Table~\ref{tab:grice-maxims}, Grice characterizes effective communication in terms of four maxims---\textit{Quantity}, \textit{Quality,} \textit{Relation}, and \textit{Manner}---that regulate how speakers cooperate in interaction \citep{grice1975logic}. 

\begin{table}[h]
\centering
\small
\setlength{\tabcolsep}{4pt}
\renewcommand{\arraystretch}{1.1}
\begin{tabularx}{\columnwidth}{l X}
\toprule
\textbf{Maxim} & \textbf{Description} \\
\midrule
\textit{Quantity} &
Make your contribution as informative as required for the current conversational goal, but not more informative than necessary. \\

\textit{Quality} &
Ensure that your contribution is truthful: do not say what you believe to be false or lack adequate evidence for. \\

\textit{Relation} &
Make your contribution relevant to the conversational context and the question under discussion. \\

\textit{Manner} &
Be clear and orderly in how you express your contribution; avoid unnecessary ambiguity. \\
\bottomrule
\end{tabularx}
\caption{The four Gricean maxims that underlie our pragmatic counterfactual taxonomy.}
\label{tab:grice-maxims}
\end{table}

\begin{table*}[!ht]
\centering
\small
\setlength{\tabcolsep}{4pt}
\renewcommand{\arraystretch}{1.1}
\begin{tabularx}{\textwidth}{l l X X}
\toprule
\textbf{Counterfactual Type} & \textbf{Pragmatic Maxim} &
\textbf{Linguistic Motivation} &
\textbf{Example} \\
\midrule
Presupposition Flip & \textit{Quality + Relation} &
Test truthfulness of assumed background knowledge; revise false presuppositions &
$Q$: Did the king of France visit Berlin?\newline
$Q'$: If there were no king of France but a president instead, would the president visit Berlin? \\

Lexical Substitution & \textit{Relation + Quantity} &
Check whether changes in lexical strength affect inference relevance &
$Q$: Could a cheetah outrun a car?\newline
$Q'$: Could a cheetah outrun a \emph{sports car}? \\

Scalar Adjustment & \textit{Quantity} &
Examine model sensitivity to implicature scales (some $\leftrightarrow$ most $\leftrightarrow$ all) &
$Q$: Do all birds fly?\newline
$Q'$: Do \emph{some} birds fly? \\

Contextualization & \textit{Relation + Manner} &
Add missing contextual information that affects interpretation &
$Q$: Is this animal a good pet?\newline
$Q'$: Given that this animal is a large wild tiger, is it a good pet? \\
\bottomrule
\end{tabularx}
\caption{Pragmatic counterfactual question types $Q'$, their associated Gricean maxims, linguistic motivations, and illustrative examples.}
\label{tab:pragmatic-taxonomy}
\end{table*}

These maxims capture implicit expectations about how much information to provide, how truthful and well-founded it should be, how relevant it is to the current goal, and how clearly it is expressed \citep{krause-vossen-2024-gricean-maxims, ma-etal-2025-pragmatics}.  As NLP systems are increasingly used in interactive and explanation-heavy settings, it is natural to ask whether the explanations they provide remain robust when these pragmatic dimensions are different \citep{jacquet19,kaczmarek2022, Alexandris2024GenAIAS, zuo2025evadellmbasedexplanationgeneration}

We therefore take the Gricean maxims as a principled basis for designing counterfactual question types. Each counterfactual corresponds to a controlled manipulation of one or more maxims. Table~\ref{tab:pragmatic-taxonomy} summarizes our perturbation methods, together with linguistic motivations and examples.

\section{Experimental Design}

\subsection{Task and Dataset} \label{sec:dataset}
We use the \textbf{StrategyQA} \citep{geva-etal-2021-aristotle} dataset as our primary QA task. StrategyQA is an English question answering benchmark of 2,780 short, open-domain yes/no questions that require implicit factual reasoning, often illustrated by examples like \textit{Did Aristotle use a laptop?}. Each example is annotated with a decomposition into reasoning steps and a supporting Wikipedia paragraph.

\subsection{Counterfactual Question Generation} \label{sec:question-generation}

\paragraph{LLM automatic counterfactual generation.} 
Following \citet{chen2024}, we prompt strong LLMs to generate counterfactual variants of each StrategyQA question. These serve as a \textit{non-pragmatic} baseline. We adopt the same LLMs as \citet{chen2024} (GPT-3.5 \citep{gpt3, Ouyang2022TrainingLM} and GPT-4 \citep{Achiam2023GPT4TR}) to ensure comparability of results, and additionally include Llama-3.3-70B-Instruct ("Llama-3.3") \citep{llama} as a strong open-source model. The prompts used for automatic counterfactual generation are listed in Appendix~\ref{sec:llm-question-generation}.

\paragraph{Pragmatics-based counterfactual generation.}
For generating pragmatics-based counterfactual questions in \textbf{StrategyQA}, we follow the taxonomy in Table~\ref{tab:pragmatic-taxonomy}. In \textbf{StrategyQA}, each question is paired with a \texttt{term} (from which the question is derived), a natural language description of that term, and a set of implicit facts needed to answer the question \citep{geva-etal-2021-aristotle}. We treat the \texttt{term} as a \textit{pragmatic anchor}---a lexical or conceptual element that carries implicit assumptions, invites alternative interpretations, or structures the reasoning process. We then construct five types of pragmatic counterfactual transformations on top of these anchors:

\begin{itemize}[leftmargin=*]
    \item \textbf{Presupposition flip}: we prompt GPT-4 to generate counterfactual variants of the form \emph{``If A is not B but C, \dots''}, thereby flipping or challenging the presuppositions associated with the anchor term. The exact prompts are provided in Appendix~\ref{sec:presupp-prompt}.
    \item \textbf{Lexical substitution}: we replace the key \texttt{term} with a synonym or hypernym retrieved from WordNet \citep{miller-1992-wordnet}, altering the lexical realization while preserving closely related meaning.
    \item \textbf{Scalar adjustment}: we substitute quantifiers with stronger or weaker scalar alternatives, using manually specified scalar scales listed in Appendix~\ref{sec:scales}. The scales are defined on the basis of the scalar inventories discussed in previous works \citep{carston98, papafragou_scalar_2003,sauerland_scalar_2004}.
    \item \textbf{Contextualization}: we append one or two supporting facts from the dataset to enrich the question context and make part of the originally implicit pragmatic content explicit.
\end{itemize}

\subsection{Self-explanations Collection}

For each original question $Q$ and its model answer $A$, we collect two types of explanations from the same set of models used for LLM automatic counterfactual generation in \S\ref{sec:question-generation}. In our study, we consider two types of verbalized self-explanations and generate one of each type for every instance in the dataset (see Appendix~\ref{sec:explanation-generator} for prompt templates):

\paragraph{Chain-of-thought (CoT) explanations}, where the model reveals an explicit step-by-step reasoning process \citep{Nye2021ShowYW}. We elicit CoT explanations using \textit{Let's think step by step} \citep{kojima2022} together with additional instructions about formatting the final answer.

\paragraph{Post-hoc explanations}, where the model is asked to justify its produced answer without modifying it \citep{NEURIPS2018_4c7a167b, park2018}.

\subsection{Counterfactual Simulation}\label{tab:simulation}

\begin{figure*}[t]
  \includegraphics[width=\textwidth]{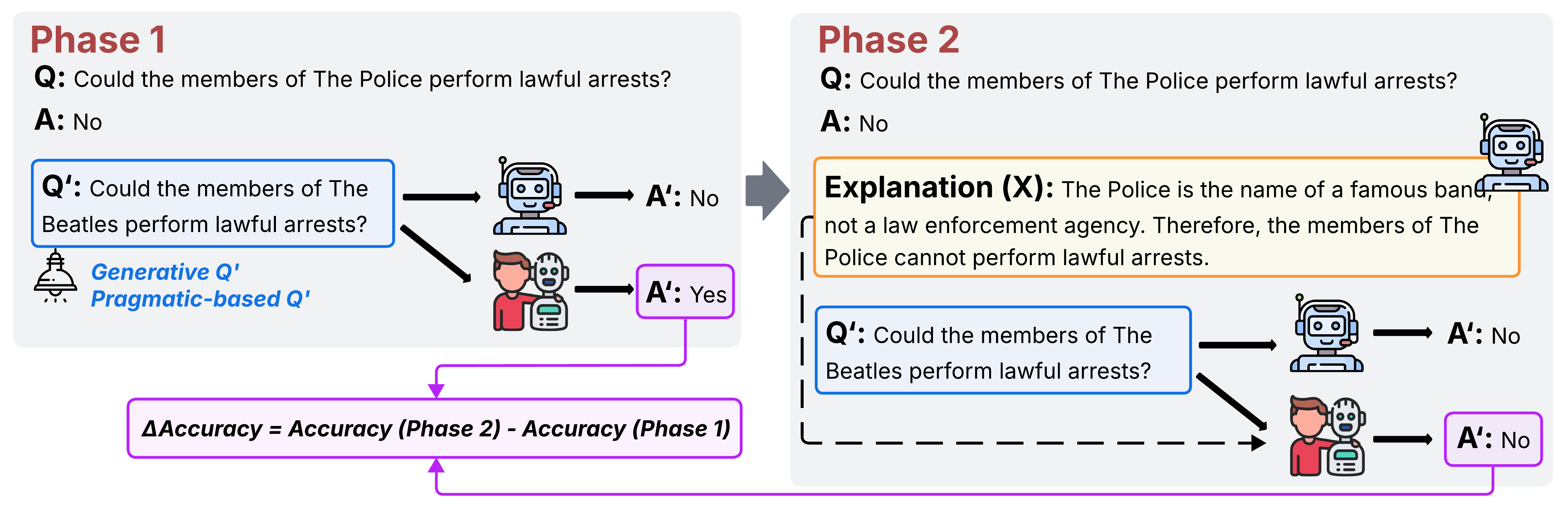}
  \caption{Counterfactual simulation pipeline for evaluating how human and LLM judges predict model behavior from explanations and counterfactual question sets.}
  \label{fig:simulation}
\end{figure*}

We use a counterfactual simulation test to measure the ability of human participants and LLM simulators to predict model behavior, following the setup of \citet{hase-bansal-2020-evaluating} and \citet{chen2024}. Our counterfactual simulation design is illustrated in Figure~\ref{fig:simulation}. In previous work, counterfactual simulation procedures have been used to assess explanation usefulness by testing whether an explanation enables a judge to infer a model’s outputs on counterfactual variants of the input. 

Our primary summary statistic is \textbf{acc\_improvement}, which we use to compare conditions. Intuitively, \textbf{OverallAcc} measures how well an evaluator can predict the model’s follow-up answer within a phase, while \textbf{acc\_improvement} captures the additional benefit of providing the model’s self-explanation by comparing Phase~2 (with explanation) against Phase~1 (without explanation). We define:
\[
\Delta\mathrm{OverallAcc}
= \mathrm{OverallAcc}^{\text{P2}} - \mathrm{OverallAcc}^{\text{P1}} .
\]
A positive $\Delta\mathrm{OverallAcc}$ indicates that access to explanations improves model-behavior prediction, whereas values near zero suggest limited impact (e.g., when Phase~1 accuracy is already high), and negative values indicate that explanations can occasionally mislead the evaluator.

In our setup, we additionally use the simulation test to study the effect of different 
counterfactual question sets. Holding the explanation set fixed, we ask which counterfactual set 
yields the largest \textbf{acc\_improvement}.

\paragraph{LLM-as-a-judge simulation.}
Building on prior work showing that LLMs approximate human judgments with high agreement and nuance, we adopt an \textit{LLM-as-a-judge} setup for our counterfactual study \citep{zheng_judge,bai2023,li-etal-2025-generation}. For each item, we pair a starter question and the model’s answer with a counterfactual follow-up. The judge then decides whether the starter answer, optionally with its explanation, allows inferring the model’s answer to the follow-up. To test robustness, we follow the scalable-oversight setup of \citet{kenton2024} and use two independent judges with different capacities as \textbf{Weaker Evaluator} and \textbf{Stronger Evaluator}.

\paragraph{Human simulation user study.} \label{sec: user-study}
To assess how well our LLM-as-a-judge evaluations align with real human behavior, we conduct a human-simulation user study on a subset of GPT-4–generated explanations (both CoT and post-hoc), since it is the most advanced explainer among the three models considered. We select 50 items, each paired with 7 counterfactual questions using 7 different perturbation methods. In each trial, participants see a scenario involving a “robot” (a LLM): a yes/no \textit{Starter Question}, the \textit{Robot’s Answer}, and a yes/no \textit{Follow-up Question} in Phase 1; in Phase 2, they additionally see a textual \textit{Explanation} of the robot’s reasoning. Participants are explicitly instructed to predict the robot’s answer rather than state their own belief about the correct answer. For every trial, they (i) rate how confident they are about the robot’s follow-up answer on a 1–5 scale, (ii) make a guess about the robot’s answer, and (iii) provide a short free-text justification of how they arrived at this guess. These annotations allow us to analyze both the accuracy and perceived difficulty of simulating the model’s behavior, as well as the role of explanations. Detailed annotation guidelines are provided in the Appendix~\ref{sec:user-guideline}.

\section{LLM-as-a-Judge Results}\label{sec:results}

\subsection{LLM Judge Performance: Analysis of Accuracy Improvements}

\begin{figure*}[t]
  \centering
  \includegraphics[width=0.85\textwidth]{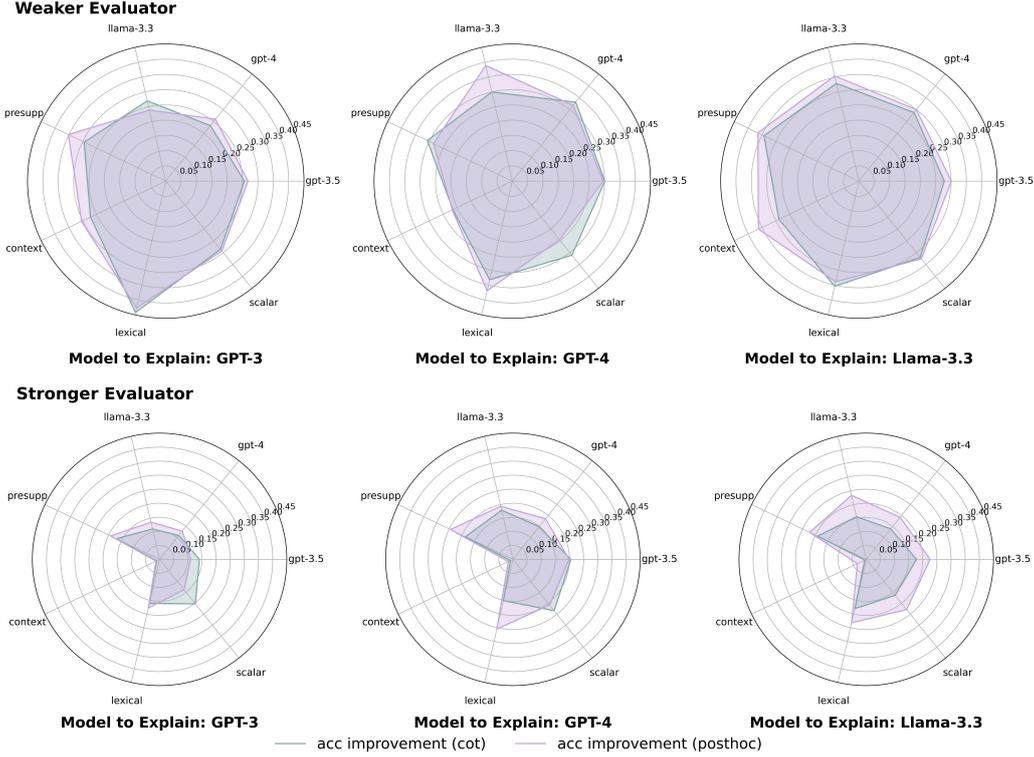}
  \caption{Radar charts summarizing LLM-as-a-judge \textbf{acc\_improvement} on different counterfactual question sets. The upper and lower panels show results for a weaker and a stronger evaluator, respectively, each using combinations of chain-of-thought (cot) and post-hoc explanations for three underlying models to be explained.}
  \label{fig:radar}
\end{figure*}

LLM-as-judge \textbf{acc\_improvement} across different counterfactual question sets is visualized with radar charts in Figure~\ref{fig:radar}. Overall, three trends emerge:

\textbf{(2) Comparison across counterfactual generation methods.}
For the weaker judge, \textit{Lexical Substitution} and \textit{Presupposition Flip} yield the most competitive results: \textit{Lexical Substitution} achieves the largest \textbf{acc\_improvement} in four out of the six explanation settings, and \textit{Presupposition Flip} also consistently performs well. Some LLM-generated counterfactual sets can be competitive in specific cases (e.g., for post-hoc explanations of GPT-4), but their gains are not stable across models and explanation types. For the stronger judge, the trends change. \textit{Scalar Adjustment} becomes one of the best-performing and most stable question type, and \textit{Presupposition Flip} remains strong except for the CoT explanations by Llama-3.3. By contrast, the effectiveness of \textit{Contextualization} drops substantially compared to the weaker judge: for the stronger judge, the Phase 1 accuracy on contextualized questions already reaches 0.627 (the highest among all counterfactual sets and much higher than the weaker judge’s 0.405), leaving very limited room for further gains in Phase 2. Overall, while no single counterfactual recipe dominates in all settings, pragmatics-based counterfactual questions tend to be more stable than purely generative ones, particularly \textit{Lexical Substitution}, \textit{Presupposition Flip}, and \textit{Scalar Adjustment}.

\textbf{(3) CoT explanations vs.\ post-hoc explanations.}
We do not observe a clear advantage of CoT explanations over post-hoc rationales. On average, post-hoc explanations perform slightly better: for the weaker judge, mean \textbf{acc\_improvement} is 0.302 for CoT vs.\ 0.315 for post-hoc, and for the stronger judge it is 0.149 vs.\ 0.156, respectively. This suggests that, in our setup, targeted rationales are already sufficient for enabling LLM judges to simulate the model’s behavior, and richer step-by-step reasoning does not translate into larger gains.

\textbf{(4) Effect of judge strength.}
Finally, comparing the two rows of radar plots shows that overall \textbf{acc\_improvement} is substantially larger for the weaker evaluator than for the stronger one: the areas of the polygons are visibly bigger in the weaker-judge row. This indicates that the same combination of explanations and counterfactual questions yields much larger gains when the judge is weaker, whereas a stronger judge already predicts the base model reasonably well in Phase 1 and therefore has limited headroom to benefit from additional information. In other words, the marginal utility of explanations and counterfactuals decreases as the judge becomes stronger.

Complete per-condition tables reporting \textbf{guess rate}, \textbf{selective accuracy}, and \textbf{overall accuracy} by phase and condition are provided in Appendix~\ref{sec:full-judge-results}.

\subsection{In-context Bias and Flip-back Behavior}
A common failure mode of LLM-judges is \emph{anchoring}: when asked to predict the model's answer to a follow-up question, the judge may simply reuse the model's original answer to the starter question, regardless of whether that original answer is correct. Here we quantify this anchoring effect and test whether providing self-explanations mitigates it, especially in cases where the model's original answer is wrong (where \emph{flip-back} is desirable).

For each instance \(i\), \(Q_i\) and \(Q'_i\) denote the starter and follow-up questions. The model's answer to \(Q_i\) is \(a_i \in \{\text{yes}, \text{no}\}\). The judge predicts the model's answer to the follow-up question as \(\hat{y}_i \in \{\text{yes}, \text{no}\}\), and we denote the phase by \(p_i \in \{1,2\}\). 

\paragraph{Anchoring to the original answer (overall).} We first measure how often the judge repeats the model's original answer when predicting the answer to \(Q'_i\). We define an indicator that the judge's prediction matches the original answer: \(B_i = \mathbb{I}[\hat{y}_i = a_i]\). \(B_i = 1\) if the judge's prediction for \(Q'_i\) exactly equals the model's answer to \(Q_i\).

For each phase \(p \in \{1,2\}\), we define the \textit{}{in-context bias toward the original answer} as:

\begin{align*}
    \widehat{\operatorname{bias}}^{\text{orig}}(p)
    &= \frac{1}{N_p} \sum_{i:\, p_i = p} B_i \, ,
\end{align*}
where \(N_p\) is the number of instances in phase \(p\).

\paragraph{Anchoring when the model is wrong (flip-back regime).} To isolate cases where anchoring is clearly undesirable, we focus on instances where the model is \emph{incorrect} on the starter question.
We define an indicator of whether the model is correct on the starter question: \(C_i = \mathbb{I}[a_i = g_i]\), so that \(C_i = 0\) marks instances where the model's starter answer is incorrect. Within this subset (\(C_i = 0\)), we reuse the indicator \(B_i\) and define:
\begin{align*}
    \widehat{\operatorname{bias}}^{\text{orig}}_{\text{wrong}}(p)
    &= \frac{1}{N_p^{\text{wrong}}}
       \sum_{i:\, p_i = p,\; C_i = 0} B_i \, ,
\end{align*}
where \(N_p^{\text{wrong}}\) is the number of instances in phase \(p\) for which the model's answer to the starter question is incorrect (\(C_i = 0\)). 

We compare \(\widehat{\operatorname{bias}}^{\text{orig}}(p)\) and \(\widehat{\operatorname{bias}}^{\text{orig}}_{\text{wrong}}(p)\) across phases:
\(\Delta\text{bias\_orig}=\widehat{\operatorname{bias}}^{\text{orig}}(2)-\widehat{\operatorname{bias}}^{\text{orig}}(1)\) and
\(\Delta\text{bias\_orig\_wrong}=\widehat{\operatorname{bias}}^{\text{orig}}_{\text{wrong}}(2)-\widehat{\operatorname{bias}}^{\text{orig}}_{\text{wrong}}(1)\).
Negative values indicate reduced anchoring to the original answer in Phase~2, while positive values indicate increased anchoring.
Table~\ref{tab:bias-gpt4} reports these changes for GPT-4 explanations across counterfactual question sets, both overall and restricted to instances where the model's starter answer is incorrect.
Full results are provided in Appendix~\ref{sec:results-bias-flipback}.

\begin{table}[t]
\centering
\resizebox{\columnwidth}{!}{%
\begin{tabular}{llrr}
\toprule
\textbf{expl\_type} & \textbf{cfq\_type} & \textbf{$\Delta$bias\_orig} & \textbf{$\Delta$bias\_orig\_wrong} \\
\midrule
\multicolumn{4}{c}{\cellcolor[HTML]{EFEFEF}\textit{\textbf{Weaker Evaluator}}} \\
cot     & gpt-3.5    & -0.013 & -0.209 \\
cot     & gpt-4            &  \cellcolor{blue!15}0.009 & -0.224 \\
cot     & Llama-3.3        & -0.043 & -0.229 \\
cot     & presupp\_flip & -0.195 & -0.017 \\
cot     & contextual       & -0.003 & -0.047 \\
cot     & lexical          & -0.057 & -0.073 \\
cot     & scalar           & -0.074 & -0.076 \\
\midrule
posthoc & gpt-3.5    & -0.029 & -0.241 \\
posthoc & gpt-4            & -0.033 & -0.304 \\
posthoc & Llama-3.3        & -0.081 & -0.284 \\
posthoc & presupp\_flip & -0.237 &  \cellcolor{blue!15}0.013 \\
posthoc & contextual       &  \cellcolor{blue!15}0.008 & -0.026 \\
posthoc & lexical          & -0.054 & -0.095 \\
posthoc & scalar           & -0.027 & -0.042 \\
\multicolumn{4}{c}{\cellcolor[HTML]{EFEFEF}\textit{\textbf{Stronger Evaluator}}} \\
cot     & gpt-3.5    & -0.090 & -0.241 \\
cot     & gpt-4            &  -0.057 & -0.201 \\
cot     & Llama-3.3        & -0.074 & -0.225 \\
cot     & presupp\_flip & -0.073 & -0.014 \\
cot     & contextual       & -0.017 & -0.084 \\
cot     & lexical          & -0.061 & -0.073 \\
cot     & scalar           & -0.049 & -0.010 \\
\midrule
posthoc & gpt-3.5    & -0.091 & -0.254 \\
posthoc & gpt-4            & -0.081 & -0.295 \\
posthoc & Llama-3.3        & -0.096 & -0.290 \\
posthoc & presupp\_flip & -0.081 & -0.022 \\
posthoc & contextual       & -0.028 & -0.104 \\
posthoc & lexical          & -0.095 & -0.149 \\
posthoc & scalar           & -0.052 & -0.079 \\
\bottomrule
\end{tabular}
}
\caption{Change in in-context bias from Phase~1 to Phase~2 for explanations of GPT-4 across counterfactual question sets. Negative values indicate reduced bias. Positive numbers are marked in blue.}
\label{tab:bias-gpt4}
\end{table}

Overall, in-context bias tends to decrease from Phase~1 to Phase~2. 
This pattern suggests that LLM verbalized self-explanations generally reduce judges' tendency to stick with the original model prediction. Moreover, the reductions are particularly pronounced when conditioning on instances where the original answer is wrong, indicating that explanations are especially helpful for correcting initially incorrect model outputs.

\section{Human User-study Results}

\subsection{Confidence and Accuracy Gains}

Under the setup in \S\ref{tab:simulation}, the mean confidence (1–5) from 28 participants increases from \textbf{3.21} without explanations to \textbf{4.38} with explanations, suggesting that explanations improve confidence in predicting the LLM’s (GPT-4) behavior.

\begin{table}[h]
\centering
\resizebox{0.8\columnwidth}{!}{
\begin{tabular}{lccc}
\toprule
\textbf{cfq\_type} & P1\_acc & P2\_acc & \textbf{$\Delta$ acc} \\
\midrule
gpt-3.5          & 0.560 & 0.540 & -0.020 \\
gpt-4          & 0.460 & 0.470 & 0.010 \\
Llama-3.3          & 0.580 & 0.580 & 0.000 \\
presupp\_flip        & 0.360 & 0.400 & 0.040 \\
contextual & 0.760 & 0.820 & \textbf{0.060} \\
lexical          & 0.760 & 0.800 & 0.040 \\
scalar         & 0.820 & 0.840 & 0.020 \\
\midrule
\textbf{Overall} & 0.614 & 0.634 & 0.020 \\
\bottomrule
\end{tabular}
}
\caption{Human user study evaluation results. Accuracy by counterfactual question type for Phase 1 and 2. For each counterfactual type, we evaluate 100 questions.}
\label{tab:user-study-acc}
\end{table}

In Table~\ref{tab:user-study-acc}, we report human prediction accuracy against the model’s actual follow-up answers in Phase 1 and Phase 2, along with the change in accuracy ($\Delta$ acc). Overall accuracy increases slightly from 0.614 to 0.634. The gain is type-dependent: \textit{Contextualization} shows the largest improvement (+0.060), while \textit{Lexical Substitution} and \textit{Presupposition Flip} improve by +0.040. In contrast, \textit{Scalar Adjustment} changes only marginally (+0.020), and model-generated types show little or no benefit, with a small drop for GPT-3.5. Overall, explanations provide modest and uneven improvements in human ability to anticipate the model’s behavior.

By comparing human users to LLM-as-a-judge results, we estimate how well LLMs approximate human judgments. On the subset that humans annotated, the LLM judge matches human yes/no predictions at an overall rate of 0.601 in Phase~1 and 0.591 in Phase~2. We provide a more detailed breakdown by counterfactual type in the Appendix~\ref{sec:appendix-human-llm-agreement}.

\subsection{Case Study: Analysis of Human Rationales for Predicting Model Behavior} \label{sec:human-explanations}
In addition to accuracy and agreement metrics, we collected participants' free-text rationales describing how they predicted the model's follow-up answer. We use these rationales to qualitatively examine what cues humans attend to when simulating model behavior, and how those cues shift from Phase~1 to Phase~2. In this section, we present a small set of representative case studies selected to reflect the main quantitative patterns.

\paragraph{Case 1: Explanations can help when they state a decision criterion that the model also applies.}
When the follow-up question directly refers to a concept mentioned in the model’s explanation, participants often move from simple Phase~1 heuristics (e.g., reusing the starter answer) to \textit{criterion matching} in Phase~2. In these cases, their rationales typically quote or paraphrase a key phrase from the explanation and apply it to the follow-up question. 

Example:
\textcolor{MidnightBlue}{\textit{In the explanation, we know that there are programs like the National School Lunch Program that provide free or reduced-price lunches to eligible students. So few students are guaranteed lunch at school in the US.}}

\paragraph{Case 2: Limited gains under ceiling effects.}
When the follow-up question is a near-paraphrase of the starter question or introduces only a minor adjustment, many participants already succeed in Phase~1 by applying a simple consistency heuristic (predicting that the model will answer the same way). In these cases, rationales often acknowledge that the explanation is consistent with their initial guess but does not add further information. Once Phase~1 performance is near ceiling, explanations have little room to improve correctness.

Example:
\textcolor{MidnightBlue}{\textit{The explanation further support the answer with further information like "The Smithsonian’s National Zoo in Washington DC is a place where various animals are kept, but it does not currently house harbor seals."}}

\paragraph{Case 3: Explanations may not resolve condition flips.}
In these cases, participants’ rationales often express uncertainty about how the model will respond after the follow-up introduces a critical change in conditions. Even in Phase~2, the model’s explanation may not directly address the specific condition that is altered in the follow-up, leaving participants without a stable rule to extrapolate the model’s behavior. As a result, these items can remain difficult across phases. Qualitatively, the rationales suggest that participants focus on the condition shift in the follow-up, while the explanation highlights other aspects, which makes it less useful for predicting the model’s answer.

Example:
\textcolor{MidnightBlue}{\textit{The explanation states that "if all conditions align", but in the follow-up question, the condition changes. Not sure whether the robot still thinks that a dealer could buy a Boeing 737-800.}}

\paragraph{Case 4: Explanations can distract when they are persuasive but not diagnostic.}
We observe cases where a participant's Phase~1 prediction is correct under a simple heuristic, but Phase~2 introduces over-reliance on a plausible-sounding detail in the explanation that does not actually determine the model's follow-up answer. In these instances, the explanation functions as a \textit{persuasive narrative} rather than a diagnostic cue, and participants may flip from a correct to an incorrect prediction. 

Example:
\textcolor{MidnightBlue}{\textit{The explanation already compares mail carriers to more dangerous jobs, so police officers will be judged dangerous.}}

\paragraph{Summary.}
Across these case studies, human rationales suggest two recurring themes. First, participants are successful when they can align the follow-up question to an explicit criterion stated in the explanation; this is precisely when Phase~2 yields the largest gains. Second, when follow-up questions are either trivial (ceiling effects) or underspecified (insufficient cues), explanations offer limited additional benefit and may even distract.

\section{Conclusion}

In this work, we assess LLM self-explanations through counterfactual simulatability. Explanations improve users’ ability to anticipate model behavior for both human participants and LLM judges, but the measured effect depends on the counterfactual construction and the evaluator. We further observe that explanations mitigate a tendency to echo the model’s original answer—most notably when the initial answer is incorrect—suggesting a corrective “flip-back” mechanism. In the human study, explanations also increase confidence and yield measurable accuracy gains.

Overall, our findings show that, despite concerns about faithfulness of verbalized self-explanations, they can help users better understand model behavior in counterfactual settings. Our results highlight that the counterfactual questions themselves are a key part of the evaluation: different perturbation strategies can change both task difficulty and the benefit of explanations, so counterfactual design should be treated as a component rather than a neutral choice. Annotations, explanations, and code will be released publicly upon publication.

\section*{Limitations}
This study has potential limitations. Our study measures model understanding using counterfactual simulatability on binary questions. While this design offers a controlled way to measure whether explanations help users predict model behavior, it may not fully reflect real-world interactions where users ask open-ended questions and engage in multi-turn dialogue. Second, although we introduce a pragmatics-grounded taxonomy of counterfactual perturbations, the set of transformations is incomplete and may not cover pragmatic phenomena that arise in natural conversations (e.g., discourse context, speaker intent, social norms).

Thirdly, parts of our pipeline rely on LLM-generated artifacts (e.g., LLM-generated counterfactuals and explanations, and GPT-4–based presupposition flips), such as counterfactual questions or explanations. Results may depend on the chosen prompts and models. We also test a limited number of models and explanation formats, so generalization is not guaranteed.

Finally, the human user study uses a relatively small sample (participants and items) and focuses on a subset of conditions, which limits statistical power for small effects and may not generalize across user populations or expertise levels. Importantly, our work does not claim that self-explanations are faithful accounts of the model’s internal decision process: explanations can improve simulatability while still being post-hoc or persuasive rather than diagnostic.

\section*{Acknowledgments} 

We thank the members of the University of Vienna Natural Language Processing (NLP) working group for their insightful feedback on earlier drafts of this paper. We also express our sincere gratitude to the 28 volunteer annotators who participated in the user study.

This research has been funded by the Vienna Science and Technology Fund (WWTF)[10.47379/VRG19008] “Knowledge-infused Deep Learning for Natural Language Processing” and by the Vienna Science and Technology Fund (WWTF)[10.47379/VRG23007] “Understanding Language in Context”.

\paragraph{Ethical considerations.} We do not foresee any ethical concerns associated with this work. All analyses were conducted using publicly available datasets and models. No private or sensitive information was used. Additionally, we release our code, prompts, and documentation to support transparency and reproducibility.

\paragraph{Use of AI Assistants.} The authors acknowledge the use of ChatGPT for correcting grammatical errors, enhancing the coherence of the final manuscripts, and providing coding assistance.



\bibliography{custom}

@inproceedings{Alexandris2024GenAIAS,
  title={GenAI and Socially Responsible AI in Natural Language Processing Applications: A Linguistic Perspective},
  author={Christina Alexandris},
  booktitle={AAAI Spring Symposia},
  year={2024},
  url={https://api.semanticscholar.org/CorpusID:269971491}
}

@inproceedings{AlvarezMelis2018TowardsRI,
author = {Alvarez-Melis, David and Jaakkola, Tommi S.},
title = {Towards robust interpretability with self-explaining neural networks},
year = {2018},
publisher = {Curran Associates Inc.},
address = {Red Hook, NY, USA},
booktitle = {Proceedings of the 32nd International Conference on Neural Information Processing Systems},
pages = {7786–7795},
numpages = {10},
location = {Montr\'{e}al, Canada},
series = {NIPS'18},
url = {https://proceedings.neurips.cc/paper_files/paper/2018/file/3e9f0fc9b2f89e043bc6233994dfcf76-Paper.pdf}
}

@inproceedings{atanasova-etal-2023-faithfulness,
    title = "Faithfulness Tests for Natural Language Explanations",
    author = "Atanasova, Pepa  and
      Camburu, Oana-Maria  and
      Lioma, Christina  and
      Lukasiewicz, Thomas  and
      Simonsen, Jakob Grue  and
      Augenstein, Isabelle",
    editor = "Rogers, Anna  and
      Boyd-Graber, Jordan  and
      Okazaki, Naoaki",
    booktitle = "Proceedings of the 61st Annual Meeting of the Association for Computational Linguistics (Volume 2: Short Papers)",
    month = jul,
    year = "2023",
    address = "Toronto, Canada",
    publisher = "Association for Computational Linguistics",
    url = "https://aclanthology.org/2023.acl-short.25/",
    doi = "10.18653/v1/2023.acl-short.25",
    pages = "283--294"
}

@inproceedings{bai2023,
author = {Bai, Yushi and Ying, Jiahao and Cao, Yixin and Lv, Xin and He, Yuze and Wang, Xiaozhi and Yu, Jifan and Zeng, Kaisheng and Xiao, Yijia and Lyu, Haozhe and Zhang, Jiayin and Li, Juanzi and Hou, Lei},
title = {Benchmarking foundation models with language-model-as-an-examiner},
year = {2023},
publisher = {Curran Associates Inc.},
address = {Red Hook, NY, USA},
booktitle = {Proceedings of the 37th International Conference on Neural Information Processing Systems},
articleno = {3414},
numpages = {26},
location = {New Orleans, LA, USA},
series = {NIPS '23},
url = {https://proceedings.neurips.cc/paper_files/paper/2023/file/f64e55d03e2fe61aa4114e49cb654acb-Paper-Datasets_and_Benchmarks.pdf}
}

@inproceedings{Bhattacharjee2023TowardsLC,
  title={Towards LLM-guided Causal Explainability for Black-box Text Classifiers},
  author={Amrita Bhattacharjee and Raha Moraffah and Joshua Garland and Huan Liu},
  year={2023},
  url={https://api.semanticscholar.org/CorpusID:262459118}
}

@inproceedings{gpt3,
author = {Brown, Tom B. and Mann, Benjamin and Ryder, Nick and Subbiah, Melanie and Kaplan, Jared and Dhariwal, Prafulla and Neelakantan, Arvind and Shyam, Pranav and Sastry, Girish and Askell, Amanda and Agarwal, Sandhini and Herbert-Voss, Ariel and Krueger, Gretchen and Henighan, Tom and Child, Rewon and Ramesh, Aditya and Ziegler, Daniel M. and Wu, Jeffrey and Winter, Clemens and Hesse, Christopher and Chen, Mark and Sigler, Eric and Litwin, Mateusz and Gray, Scott and Chess, Benjamin and Clark, Jack and Berner, Christopher and McCandlish, Sam and Radford, Alec and Sutskever, Ilya and Amodei, Dario},
title = {Language models are few-shot learners},
year = {2020},
isbn = {9781713829546},
publisher = {Curran Associates Inc.},
address = {Red Hook, NY, USA},
booktitle = {Proceedings of the 34th International Conference on Neural Information Processing Systems},
articleno = {159},
numpages = {25},
location = {Vancouver, BC, Canada},
series = {NIPS '20},
url = {https://papers.nips.cc/paper/2020/hash/1457c0d6bfcb4967418bfb8ac142f64a-Abstract.html}
}

@inproceedings{NEURIPS2018_4c7a167b,
 author = {Camburu, Oana-Maria and Rockt\"{a}schel, Tim and Lukasiewicz, Thomas and Blunsom, Phil},
 booktitle = {Advances in Neural Information Processing Systems},
 editor = {S. Bengio and H. Wallach and H. Larochelle and K. Grauman and N. Cesa-Bianchi and R. Garnett},
 pages = {},
 publisher = {Curran Associates, Inc.},
 title = {e-SNLI: Natural Language Inference with Natural Language Explanations},
 url = {https://proceedings.neurips.cc/paper_files/paper/2018/file/4c7a167bb329bd92580a99ce422d6fa6-Paper.pdf},
 volume = {31},
 year = {2018}
}

@article{carston98,
author = {Carston, Robyn},
year = {1998},
month = {01},
pages = {},
title = {Informativeness, Relevance and Scalar Implicature},
journal = {Relevance Theory. Applications and Implications},
doi = {10.1075/pbns.37.11car}
}

@inproceedings{chandrasekaran-etal-2018-explanations,
    title = "Do explanations make {VQA} models more predictable to a human?",
    author = "Chandrasekaran, Arjun  and
      Prabhu, Viraj  and
      Yadav, Deshraj  and
      Chattopadhyay, Prithvijit  and
      Parikh, Devi",
    editor = "Riloff, Ellen  and
      Chiang, David  and
      Hockenmaier, Julia  and
      Tsujii, Jun{'}ichi",
    booktitle = "Proceedings of the 2018 Conference on Empirical Methods in Natural Language Processing",
    month = oct # "-" # nov,
    year = "2018",
    address = "Brussels, Belgium",
    publisher = "Association for Computational Linguistics",
    url = "https://aclanthology.org/D18-1128/",
    doi = "10.18653/v1/D18-1128",
    pages = "1036--1042"
}

@inproceedings{chen2024,
author = {Chen, Yanda and Zhong, Ruiqi and Ri, Narutatsu and Zhao, Chen and He, He and Steinhardt, Jacob and Yu, Zhou and McKeown, Kathleen},
title = {Do models explain themselves? counterfactual simulatability of natural language explanations},
year = {2024},
publisher = {JMLR.org},
booktitle = {Proceedings of the 41st International Conference on Machine Learning},
articleno = {310},
numpages = {25},
location = {Vienna, Austria},
series = {ICML'24},
url = {https://dl.acm.org/doi/10.5555/3692070.3692380}
}

@article{DoshiVelez2017TowardsAR,
  title={Towards A Rigorous Science of Interpretable Machine Learning},
  author={Finale Doshi-Velez and Been Kim},
  journal={arXiv: Machine Learning},
  year={2017},
  url={https://api.semanticscholar.org/CorpusID:11319376}
}

@inproceedings{gardner-etal-2020-evaluating,
    title = "Evaluating Models' Local Decision Boundaries via Contrast Sets",
    author = "Gardner, Matt  and
      Artzi, Yoav  and
      Basmov, Victoria  and
      Berant, Jonathan  and
      Bogin, Ben  and
      Chen, Sihao  and
      Dasigi, Pradeep  and
      Dua, Dheeru  and
      Elazar, Yanai  and
      Gottumukkala, Ananth  and
      Gupta, Nitish  and
      Hajishirzi, Hannaneh  and
      Ilharco, Gabriel  and
      Khashabi, Daniel  and
      Lin, Kevin  and
      Liu, Jiangming  and
      Liu, Nelson F.  and
      Mulcaire, Phoebe  and
      Ning, Qiang  and
      Singh, Sameer  and
      Smith, Noah A.  and
      Subramanian, Sanjay  and
      Tsarfaty, Reut  and
      Wallace, Eric  and
      Zhang, Ally  and
      Zhou, Ben",
    editor = "Cohn, Trevor  and
      He, Yulan  and
      Liu, Yang",
    booktitle = "Findings of the Association for Computational Linguistics: EMNLP 2020",
    month = nov,
    year = "2020",
    address = "Online",
    publisher = "Association for Computational Linguistics",
    url = "https://aclanthology.org/2020.findings-emnlp.117/",
    doi = "10.18653/v1/2020.findings-emnlp.117",
    pages = "1307--1323"
}

@article{Ge2021CounterfactualEF,
  author       = {Yingqiang Ge and
                  Shuchang Liu and
                  Zelong Li and
                  Shuyuan Xu and
                  Shijie Geng and
                  Yunqi Li and
                  Juntao Tan and
                  Fei Sun and
                  Yongfeng Zhang},
  title        = {Counterfactual Evaluation for Explainable {AI}},
  journal      = {CoRR},
  volume       = {abs/2109.01962},
  year         = {2021},
  url          = {https://arxiv.org/abs/2109.01962},
  eprinttype    = {arXiv},
  eprint       = {2109.01962},
  bibsource    = {dblp computer science bibliography, https://dblp.org}
}

@article{geva-etal-2021-aristotle,
    title = "Did Aristotle Use a Laptop? A Question Answering Benchmark with Implicit Reasoning Strategies",
    author = "Geva, Mor  and
      Khashabi, Daniel  and
      Segal, Elad  and
      Khot, Tushar  and
      Roth, Dan  and
      Berant, Jonathan",
    editor = "Roark, Brian  and
      Nenkova, Ani",
    journal = "Transactions of the Association for Computational Linguistics",
    volume = "9",
    year = "2021",
    address = "Cambridge, MA",
    publisher = "MIT Press",
    url = "https://aclanthology.org/2021.tacl-1.21/",
    doi = "10.1162/tacl_a_00370",
    pages = "346--361"
}

@article{Gilpin2018ExplainingEA,
  title={Explaining Explanations: An Overview of Interpretability of Machine Learning},
  author={Leilani H. Gilpin and David Bau and Ben Z. Yuan and Ayesha Bajwa and Michael A. Specter and Lalana Kagal},
  journal={2018 IEEE 5th International Conference on Data Science and Advanced Analytics (DSAA)},
  year={2018},
  pages={80-89},
  url={https://api.semanticscholar.org/CorpusID:59600034}
}

@incollection{grice1975logic,
  address = {New York},
  author = {Grice, H. P.},
  booktitle = {Syntax and Semantics: Vol. 3: Speech Acts},
  description = {Daniel Sonntag all references},
  editor = {Cole, Peter and Morgan, Jerry L.},
  pages = {41-58},
  publisher = {Academic Press},
  title = {Logic and Conversation},
  url = {http://www.ucl.ac.uk/ls/studypacks/Grice-Logic.pdf},
  username = {porta},
  year = 1975
}

@inproceedings{hase-bansal-2020-evaluating,
    title = "Evaluating Explainable {AI}: Which Algorithmic Explanations Help Users Predict Model Behavior?",
    author = "Hase, Peter  and
      Bansal, Mohit",
    editor = "Jurafsky, Dan  and
      Chai, Joyce  and
      Schluter, Natalie  and
      Tetreault, Joel",
    booktitle = "Proceedings of the 58th Annual Meeting of the Association for Computational Linguistics",
    month = jul,
    year = "2020",
    address = "Online",
    publisher = "Association for Computational Linguistics",
    url = "https://aclanthology.org/2020.acl-main.491/",
    doi = "10.18653/v1/2020.acl-main.491",
    pages = "5540--5552"
}

@article{Hoffman2018MetricsFE,
  author       = {Robert R. Hoffman and
                  Shane T. Mueller and
                  Gary Klein and
                  Jordan Litman},
  title        = {Metrics for Explainable {AI:} Challenges and Prospects},
  journal      = {CoRR},
  volume       = {abs/1812.04608},
  year         = {2018},
  url          = {http://arxiv.org/abs/1812.04608},
  eprinttype    = {arXiv},
  eprint       = {1812.04608},
  bibsource    = {dblp computer science bibliography, https://dblp.org}
}

@inproceedings{Jacovi2020TowardsFI,
  title={Towards Faithfully Interpretable NLP Systems: How Should We Define and Evaluate Faithfulness?},
  author={Alon Jacovi and Yoav Goldberg},
  booktitle={Annual Meeting of the Association for Computational Linguistics},
  year={2020},
  url={https://api.semanticscholar.org/CorpusID:215416110}
}

@inbook{jacquet19,
author = {Jacquet, Baptiste and Masson, Olivier and Jamet, Frank and Baratgin, Jean},
year = {2019},
month = {10},
pages = {394-399},
title = {On the Lack of Pragmatic Processing in Artificial Conversational Agents},
isbn = {978-3-030-02052-1},
doi = {10.1007/978-3-030-02053-8_60}
}

@article{kaczmarek2022,
author = {Kaczmarek-Majer, Katarzyna and Casalino, Gabriella and Castellano, Giovanna and Dominiak, Monika and Hryniewicz, Olgierd and Kaminska, Olga and Vessio, Gennaro and Díaz-Rodríguez, Natalia},
year = {2022},
month = {10},
pages = {},
title = {PLENARY: Explaining black-box models in natural language through fuzzy linguistic summaries},
volume = {614},
journal = {Information Sciences},
doi = {10.1016/j.ins.2022.10.010}
}

@inproceedings{Kahneman1982TheSH,
  title={The simulation heuristic},
  author={Daniel Kahneman and Amos Tversky},
  year={1982},
  url={https://api.semanticscholar.org/CorpusID:142448674}
}

@inproceedings{kenton2024,
author = {Kenton, Zachary and Siegel, Noah Y. and Kram\'{a}r, J\'{a}nos and Brown-Cohen, Jonah and Albanie, Samuel and Bulian, Jannis and Agarwal, Rishabh and Lindner, David and Tang, Yunhao and Goodman, Noah D. and Shah, Rohin},
title = {On scalable oversight with weak LLMs judging strong LLMs},
year = {2024},
isbn = {9798331314385},
publisher = {Curran Associates Inc.},
address = {Red Hook, NY, USA},
booktitle = {Proceedings of the 38th International Conference on Neural Information Processing Systems},
articleno = {2395},
numpages = {48},
location = {Vancouver, BC, Canada},
series = {NIPS '24},
url = {https://proceedings.neurips.cc/paper_files/paper/2024/file/899511e37a8e01e1bd6f6f1d377cc250-Paper-Conference.pdf}
}

@inproceedings{kojima2022,
author = {Kojima, Takeshi and Gu, Shixiang Shane and Reid, Machel and Matsuo, Yutaka and Iwasawa, Yusuke},
title = {Large language models are zero-shot reasoners},
year = {2022},
isbn = {9781713871088},
publisher = {Curran Associates Inc.},
address = {Red Hook, NY, USA},
booktitle = {Proceedings of the 36th International Conference on Neural Information Processing Systems},
articleno = {1613},
numpages = {15},
location = {New Orleans, LA, USA},
series = {NIPS '22},
url = {https://dl.acm.org/doi/10.5555/3600270.3601883}
}

@inproceedings{krause-vossen-2024-gricean-maxims,
    title = "The {G}ricean Maxims in {NLP} - A Survey",
    author = "Krause, Lea  and
      Vossen, Piek T.J.M.",
    editor = "Mahamood, Saad  and
      Minh, Nguyen Le  and
      Ippolito, Daphne",
    booktitle = "Proceedings of the 17th International Natural Language Generation Conference",
    month = sep,
    year = "2024",
    address = "Tokyo, Japan",
    publisher = "Association for Computational Linguistics",
    url = "https://aclanthology.org/2024.inlg-main.39/",
    doi = "10.18653/v1/2024.inlg-main.39",
    pages = "470--485"
}

@inproceedings{li-etal-2025-generation,
    title = "From Generation to Judgment: Opportunities and Challenges of {LLM}-as-a-judge",
    author = "Li, Dawei  and
      Jiang, Bohan  and
      Huang, Liangjie  and
      Beigi, Alimohammad  and
      Zhao, Chengshuai  and
      Tan, Zhen  and
      Bhattacharjee, Amrita  and
      Jiang, Yuxuan  and
      Chen, Canyu  and
      Wu, Tianhao  and
      Shu, Kai  and
      Cheng, Lu  and
      Liu, Huan",
    editor = "Christodoulopoulos, Christos  and
      Chakraborty, Tanmoy  and
      Rose, Carolyn  and
      Peng, Violet",
    booktitle = "Proceedings of the 2025 Conference on Empirical Methods in Natural Language Processing",
    month = nov,
    year = "2025",
    address = "Suzhou, China",
    publisher = "Association for Computational Linguistics",
    url = "https://aclanthology.org/2025.emnlp-main.138/",
    doi = "10.18653/v1/2025.emnlp-main.138",
    pages = "2757--2791",
    ISBN = "979-8-89176-332-6"
}

@inproceedings{limpijankit-etal-2025-counterfactual,
    title = "Counterfactual Simulatability of {LLM} Explanations for Generation Tasks",
    author = "Limpijankit, Marvin  and
      Chen, Yanda  and
      Subbiah, Melanie  and
      Deas, Nicholas  and
      McKeown, Kathleen",
    editor = "Flek, Lucie  and
      Narayan, Shashi  and
      Phương, L{\^e} Hồng  and
      Pei, Jiahuan",
    booktitle = "Proceedings of the 18th International Natural Language Generation Conference",
    month = oct,
    year = "2025",
    address = "Hanoi, Vietnam",
    publisher = "Association for Computational Linguistics",
    url = "https://aclanthology.org/2025.inlg-main.38/",
    pages = "659--683"
}

@article{llama,
  author       = {Aaron Grattafiori and Abhimanyu Dubey and Abhinav Jauhri and Abhinav Pandey and Abhishek Kadian and Ahmad Al-Dahle and Aiesha Letman and Akhil Mathur and Alan Schelten and Alex Vaughan and Amy Yang and Angela Fan and Anirudh Goyal and Anthony Hartshorn and Aobo Yang and Archi Mitra and Archie Sravankumar and Artem Korenev and Arthur Hinsvark and Arun Rao and Aston Zhang and Aurelien Rodriguez and Austen Gregerson and Ava Spataru and Baptiste Roziere and Bethany Biron and Binh Tang and Bobbie Chern and Charlotte Caucheteux and Chaya Nayak and Chloe Bi and Chris Marra and Chris McConnell and Christian Keller and Christophe Touret and Chunyang Wu and Corinne Wong and Cristian Canton Ferrer and Cyrus Nikolaidis and Damien Allonsius and Daniel Song and Danielle Pintz and Danny Livshits and Danny Wyatt and David Esiobu and Dhruv Choudhary and Dhruv Mahajan and Diego Garcia-Olano and Diego Perino and Dieuwke Hupkes and Egor Lakomkin and Ehab AlBadawy and Elina Lobanova and Emily Dinan and Eric Michael Smith and Filip Radenovic and Francisco Guzmán and Frank Zhang and Gabriel Synnaeve and Gabrielle Lee and Georgia Lewis Anderson and Govind Thattai and Graeme Nail and Gregoire Mialon and Guan Pang and Guillem Cucurell and Hailey Nguyen and Hannah Korevaar and Hu Xu and Hugo Touvron and Iliyan Zarov and Imanol Arrieta Ibarra and Isabel Kloumann and Ishan Misra and Ivan Evtimov and Jack Zhang and Jade Copet and Jaewon Lee and Jan Geffert and Jana Vranes and Jason Park and Jay Mahadeokar and Jeet Shah and Jelmer van der Linde and Jennifer Billock and Jenny Hong and Jenya Lee and Jeremy Fu and Jianfeng Chi and Jianyu Huang and Jiawen Liu and Jie Wang and Jiecao Yu and Joanna Bitton and Joe Spisak and Jongsoo Park and Joseph Rocca and Joshua Johnstun and Joshua Saxe and Junteng Jia and Kalyan Vasuden Alwala and Karthik Prasad and Kartikeya Upasani and Kate Plawiak and Ke Li and Kenneth Heafield and Kevin Stone and Khalid El-Arini and Krithika Iyer and Kshitiz Malik and Kuenley Chiu and Kunal Bhalla and Kushal Lakhotia and Lauren Rantala-Yeary and Laurens van der Maaten and Lawrence Chen and Liang Tan and Liz Jenkins and Louis Martin and Lovish Madaan and Lubo Malo and Lukas Blecher and Lukas Landzaat and Luke de Oliveira and Madeline Muzzi and Mahesh Pasupuleti and Mannat Singh and Manohar Paluri and Marcin Kardas and Maria Tsimpoukelli and Mathew Oldham and Mathieu Rita and Maya Pavlova and Melanie Kambadur and Mike Lewis and Min Si and Mitesh Kumar Singh and Mona Hassan and Naman Goyal and Narjes Torabi and Nikolay Bashlykov and Nikolay Bogoychev and Niladri Chatterji and Ning Zhang and Olivier Duchenne and Onur Çelebi and Patrick Alrassy and Pengchuan Zhang and Pengwei Li and Petar Vasic and Peter Weng and Prajjwal Bhargava and Pratik Dubal and Praveen Krishnan and Punit Singh Koura and Puxin Xu and Qing He and Qingxiao Dong and Ragavan Srinivasan and Raj Ganapathy and Ramon Calderer and Ricardo Silveira Cabral and Robert Stojnic and Roberta Raileanu and Rohan Maheswari and Rohit Girdhar and Rohit Patel and Romain Sauvestre and Ronnie Polidoro and Roshan Sumbaly and Ross Taylor and Ruan Silva and Rui Hou and Rui Wang and Saghar Hosseini and Sahana Chennabasappa and Sanjay Singh and Sean Bell and Seohyun Sonia Kim and Sergey Edunov and Shaoliang Nie and Sharan Narang and Sharath Raparthy and Sheng Shen and Shengye Wan and Shruti Bhosale and Shun Zhang and Simon Vandenhende and Soumya Batra and Spencer Whitman and Sten Sootla and Stephane Collot and Suchin Gururangan and Sydney Borodinsky and Tamar Herman and Tara Fowler and Tarek Sheasha and Thomas Georgiou and Thomas Scialom and Tobias Speckbacher and Todor Mihaylov and Tong Xiao and Ujjwal Karn and Vedanuj Goswami and Vibhor Gupta and Vignesh Ramanathan and Viktor Kerkez and Vincent Gonguet and Virginie Do and Vish Vogeti and Vítor Albiero and Vladan Petrovic and Weiwei Chu and Wenhan Xiong and Wenyin Fu and Whitney Meers and Xavier Martinet and Xiaodong Wang and Xiaofang Wang and Xiaoqing Ellen Tan and Xide Xia and Xinfeng Xie and Xuchao Jia and Xuewei Wang and Yaelle Goldschlag and Yashesh Gaur and Yasmine Babaei and Yi Wen and Yiwen Song and Yuchen Zhang and Yue Li and Yuning Mao and Zacharie Delpierre Coudert and Zheng Yan and Zhengxing Chen and Zoe Papakipos and Aaditya Singh and Aayushi Srivastava and Abha Jain and Adam Kelsey and Adam Shajnfeld and Adithya Gangidi and Adolfo Victoria and Ahuva Goldstand and Ajay Menon and Ajay Sharma and Alex Boesenberg and Alexei Baevski and Allie Feinstein and Amanda Kallet and Amit Sangani and Amos Teo and Anam Yunus and Andrei Lupu and Andres Alvarado and Andrew Caples and Andrew Gu and Andrew Ho and Andrew Poulton and Andrew Ryan and Ankit Ramchandani and Annie Dong and Annie Franco and Anuj Goyal and Aparajita Saraf and Arkabandhu Chowdhury and Ashley Gabriel and Ashwin Bharambe and Assaf Eisenman and Azadeh Yazdan and Beau James and Ben Maurer and Benjamin Leonhardi and Bernie Huang and Beth Loyd and Beto De Paola and Bhargavi Paranjape and Bing Liu and Bo Wu and Boyu Ni and Braden Hancock and Bram Wasti and Brandon Spence and Brani Stojkovic and Brian Gamido and Britt Montalvo and Carl Parker and Carly Burton and Catalina Mejia and Ce Liu and Changhan Wang and Changkyu Kim and Chao Zhou and Chester Hu and Ching-Hsiang Chu and Chris Cai and Chris Tindal and Christoph Feichtenhofer and Cynthia Gao and Damon Civin and Dana Beaty and Daniel Kreymer and Daniel Li and David Adkins and David Xu and Davide Testuggine and Delia David and Devi Parikh and Diana Liskovich and Didem Foss and Dingkang Wang and Duc Le and Dustin Holland and Edward Dowling and Eissa Jamil and Elaine Montgomery and Eleonora Presani and Emily Hahn and Emily Wood and Eric-Tuan Le and Erik Brinkman and Esteban Arcaute and Evan Dunbar and Evan Smothers and Fei Sun and Felix Kreuk and Feng Tian and Filippos Kokkinos and Firat Ozgenel and Francesco Caggioni and Frank Kanayet and Frank Seide and Gabriela Medina Florez and Gabriella Schwarz and Gada Badeer and Georgia Swee and Gil Halpern and Grant Herman and Grigory Sizov and Guangyi and Zhang and Guna Lakshminarayanan and Hakan Inan and Hamid Shojanazeri and Han Zou and Hannah Wang and Hanwen Zha and Haroun Habeeb and Harrison Rudolph and Helen Suk and Henry Aspegren and Hunter Goldman and Hongyuan Zhan and Ibrahim Damlaj and Igor Molybog and Igor Tufanov and Ilias Leontiadis and Irina-Elena Veliche and Itai Gat and Jake Weissman and James Geboski and James Kohli and Janice Lam and Japhet Asher and Jean-Baptiste Gaya and Jeff Marcus and Jeff Tang and Jennifer Chan and Jenny Zhen and Jeremy Reizenstein and Jeremy Teboul and Jessica Zhong and Jian Jin and Jingyi Yang and Joe Cummings and Jon Carvill and Jon Shepard and Jonathan McPhie and Jonathan Torres and Josh Ginsburg and Junjie Wang and Kai Wu and Kam Hou U and Karan Saxena and Kartikay Khandelwal and Katayoun Zand and Kathy Matosich and Kaushik Veeraraghavan and Kelly Michelena and Keqian Li and Kiran Jagadeesh and Kun Huang and Kunal Chawla and Kyle Huang and Lailin Chen and Lakshya Garg and Lavender A and Leandro Silva and Lee Bell and Lei Zhang and Liangpeng Guo and Licheng Yu and Liron Moshkovich and Luca Wehrstedt and Madian Khabsa and Manav Avalani and Manish Bhatt and Martynas Mankus and Matan Hasson and Matthew Lennie and Matthias Reso and Maxim Groshev and Maxim Naumov and Maya Lathi and Meghan Keneally and Miao Liu and Michael L. Seltzer and Michal Valko and Michelle Restrepo and Mihir Patel and Mik Vyatskov and Mikayel Samvelyan and Mike Clark and Mike Macey and Mike Wang and Miquel Jubert Hermoso and Mo Metanat and Mohammad Rastegari and Munish Bansal and Nandhini Santhanam and Natascha Parks and Natasha White and Navyata Bawa and Nayan Singhal and Nick Egebo and Nicolas Usunier and Nikhil Mehta and Nikolay Pavlovich Laptev and Ning Dong and Norman Cheng and Oleg Chernoguz and Olivia Hart and Omkar Salpekar and Ozlem Kalinli and Parkin Kent and Parth Parekh and Paul Saab and Pavan Balaji and Pedro Rittner and Philip Bontrager and Pierre Roux and Piotr Dollar and Polina Zvyagina and Prashant Ratanchandani and Pritish Yuvraj and Qian Liang and Rachad Alao and Rachel Rodriguez and Rafi Ayub and Raghotham Murthy and Raghu Nayani and Rahul Mitra and Rangaprabhu Parthasarathy and Raymond Li and Rebekkah Hogan and Robin Battey and Rocky Wang and Russ Howes and Ruty Rinott and Sachin Mehta and Sachin Siby and Sai Jayesh Bondu and Samyak Datta and Sara Chugh and Sara Hunt and Sargun Dhillon and Sasha Sidorov and Satadru Pan and Saurabh Mahajan and Saurabh Verma and Seiji Yamamoto and Sharadh Ramaswamy and Shaun Lindsay and Shaun Lindsay and Sheng Feng and Shenghao Lin and Shengxin Cindy Zha and Shishir Patil and Shiva Shankar and Shuqiang Zhang and Shuqiang Zhang and Sinong Wang and Sneha Agarwal and Soji Sajuyigbe and Soumith Chintala and Stephanie Max and Stephen Chen and Steve Kehoe and Steve Satterfield and Sudarshan Govindaprasad and Sumit Gupta and Summer Deng and Sungmin Cho and Sunny Virk and Suraj Subramanian and Sy Choudhury and Sydney Goldman and Tal Remez and Tamar Glaser and Tamara Best and Thilo Koehler and Thomas Robinson and Tianhe Li and Tianjun Zhang and Tim Matthews and Timothy Chou and Tzook Shaked and Varun Vontimitta and Victoria Ajayi and Victoria Montanez and Vijai Mohan and Vinay Satish Kumar and Vishal Mangla and Vlad Ionescu and Vlad Poenaru and Vlad Tiberiu Mihailescu and Vladimir Ivanov and Wei Li and Wenchen Wang and Wenwen Jiang and Wes Bouaziz and Will Constable and Xiaocheng Tang and Xiaojian Wu and Xiaolan Wang and Xilun Wu and Xinbo Gao and Yaniv Kleinman and Yanjun Chen and Ye Hu and Ye Jia and Ye Qi and Yenda Li and Yilin Zhang and Ying Zhang and Yossi Adi and Youngjin Nam and Yu and Wang and Yu Zhao and Yuchen Hao and Yundi Qian and Yunlu Li and Yuzi He and Zach Rait and Zachary DeVito and Zef Rosnbrick and Zhaoduo Wen and Zhenyu Yang and Zhiwei Zhao and Zhiyu Ma},
  title        = {The Llama 3 Herd of Models},
  journal      = {CoRR},
  volume       = {abs/2407.21783},
  year         = {2024},
  url          = {https://doi.org/10.48550/arXiv.2407.21783},
  doi          = {10.48550/ARXIV.2407.21783},
  eprinttype    = {arXiv},
  eprint       = {2407.21783},
  bibsource    = {dblp computer science bibliography, https://dblp.org}
}

@inproceedings{ma-etal-2025-pragmatics,
    title = "Pragmatics in the Era of Large Language Models: A Survey on Datasets, Evaluation, Opportunities and Challenges",
    author = "Ma, Bolei  and
      Li, Yuting  and
      Zhou, Wei  and
      Gong, Ziwei  and
      Liu, Yang Janet  and
      Jasinskaja, Katja  and
      Friedrich, Annemarie  and
      Hirschberg, Julia  and
      Kreuter, Frauke  and
      Plank, Barbara",
    editor = "Che, Wanxiang  and
      Nabende, Joyce  and
      Shutova, Ekaterina  and
      Pilehvar, Mohammad Taher",
    booktitle = "Proceedings of the 63rd Annual Meeting of the Association for Computational Linguistics (Volume 1: Long Papers)",
    month = jul,
    year = "2025",
    address = "Vienna, Austria",
    publisher = "Association for Computational Linguistics",
    url = "https://aclanthology.org/2025.acl-long.425/",
    doi = "10.18653/v1/2025.acl-long.425",
    pages = "8679--8696",
    ISBN = "979-8-89176-251-0"
}

@inproceedings{Madaan2020GenerateYC,
  title={Generate Your Counterfactuals: Towards Controlled Counterfactual Generation for Text},
  author={Nishtha Madaan and Inkit Padhi and Naveen Panwar and Diptikalyan Saha},
  booktitle={AAAI Conference on Artificial Intelligence},
  year={2020},
  url={https://api.semanticscholar.org/CorpusID:228063841}
}

@inproceedings{madsen-etal-2024-self,
    title = "Are self-explanations from Large Language Models faithful?",
    author = "Madsen, Andreas  and
      Chandar, Sarath  and
      Reddy, Siva",
    editor = "Ku, Lun-Wei  and
      Martins, Andre  and
      Srikumar, Vivek",
    booktitle = "Findings of the Association for Computational Linguistics: ACL 2024",
    month = aug,
    year = "2024",
    address = "Bangkok, Thailand",
    publisher = "Association for Computational Linguistics",
    url = "https://aclanthology.org/2024.findings-acl.19/",
    doi = "10.18653/v1/2024.findings-acl.19",
    pages = "295--337"
}

@inproceedings{miller-1992-wordnet,
    title = "{W}ord{N}et: A Lexical Database for {E}nglish",
    author = "Miller, George A.",
    booktitle = "Speech and Natural Language: Proceedings of a Workshop Held at Harriman, New York, {F}ebruary 23-26, 1992",
    year = "1992",
    url = "https://aclanthology.org/H92-1116/"
}

@article{Miller2017ExplanationIA,
	title = {Explanation in artificial intelligence: {Insights} from the social sciences},
	volume = {267},
	issn = {0004-3702},
	url = {https://www.sciencedirect.com/science/article/pii/S0004370218305988},
	doi = {https://doi.org/10.1016/j.artint.2018.07.007},
	journal = {Artificial Intelligence},
	author = {Miller, Tim},
	year = {2019},
	pages = {1--38},
}

@inproceedings{nguyen-etal-2024-llms,
    title = "{LLM}s for Generating and Evaluating Counterfactuals: A Comprehensive Study",
    author = {Nguyen, Van Bach  and
      Youssef, Paul  and
      Seifert, Christin  and
      Schl{\"o}tterer, J{\"o}rg},
    editor = "Al-Onaizan, Yaser  and
      Bansal, Mohit  and
      Chen, Yun-Nung",
    booktitle = "Findings of the Association for Computational Linguistics: EMNLP 2024",
    month = nov,
    year = "2024",
    address = "Miami, Florida, USA",
    publisher = "Association for Computational Linguistics",
    url = "https://aclanthology.org/2024.findings-emnlp.870/",
    doi = "10.18653/v1/2024.findings-emnlp.870",
    pages = "14809--14824"
}

@article{Nye2021ShowYW,
  author       = {Maxwell I. Nye and
                  Anders Johan Andreassen and
                  Guy Gur{-}Ari and
                  Henryk Michalewski and
                  Jacob Austin and
                  David Bieber and
                  David Dohan and
                  Aitor Lewkowycz and
                  Maarten Bosma and
                  David Luan and
                  Charles Sutton and
                  Augustus Odena},
  title        = {Show Your Work: Scratchpads for Intermediate Computation with Language
                  Models},
  journal      = {CoRR},
  volume       = {abs/2112.00114},
  year         = {2021},
  url          = {https://arxiv.org/abs/2112.00114},
  eprinttype    = {arXiv},
  eprint       = {2112.00114},
  bibsource    = {dblp computer science bibliography, https://dblp.org}
}

@inproceedings{Achiam2023GPT4TR,
  title={GPT-4 Technical Report},
  author={OpenAI and Josh Achiam and Steven Adler and Sandhini Agarwal and Lama Ahmad and Ilge Akkaya and Florencia Leoni Aleman and Diogo Almeida and Janko Altenschmidt and Sam Altman and Shyamal Anadkat and Red Avila and Igor Babuschkin and Suchir Balaji and Valerie Balcom and Paul Baltescu and Haim-ing Bao and Mo Bavarian and Jeff Belgum and Irwan Bello and Jake Berdine and Gabriel Bernadett-Shapiro and Christopher Berner and Lenny Bogdonoff and Oleg Boiko and Made-laine Boyd and Anna-Luisa Brakman and Greg Brockman and Tim Brooks and Miles Brundage and Kevin Button and Trevor Cai and Rosie Campbell and Andrew Cann and Brittany Carey and Chelsea Carlson and Rory Carmichael and Brooke Chan and Che Chang and Fotis Chantzis and Derek Chen and Sully Chen and Ruby Chen and Jason Chen and Mark Chen and Benjamin Chess and Chester Cho and Casey Chu and Hyung Won Chung and Dave Cummings and Jeremiah Currier and Yunxing Dai and Cory Decareaux and Thomas Degry and Noah Deutsch and Damien Deville and Arka Dhar and David Dohan and Steve Dowling and Sheila Dunning and Adrien Ecoffet and Atty Eleti and Tyna Eloundou and David Farhi and Liam Fedus and Niko Felix and Sim'on Posada Fishman and Juston Forte and Is-abella Fulford and Leo Gao and Elie Georges and Christian Gibson and Vik Goel and Tarun Gogineni and Gabriel Goh and Raphael Gontijo-Lopes and Jonathan Gordon and Morgan Grafstein and Scott Gray and Ryan Greene and Joshua Gross and Shixiang Shane Gu and Yufei Guo and Chris Hallacy and Jesse Han and Jeff Harris and Yuchen He and Mike Heaton and Johannes Heidecke and Chris Hesse and Alan Hickey and Wade Hickey and Peter Hoeschele and Brandon Houghton and Kenny Hsu and Shengli Hu and Xin Hu and Joost Huizinga and Shantanu Jain and Shawn Jain and Joanne Jang and Angela Jiang and Roger Jiang and Haozhun Jin and Denny Jin and Shino Jomoto and Billie Jonn and Heewoo Jun and Tomer Kaftan and Lukasz Kaiser and Ali Kamali and Ingmar Kanitscheider and Nitish Shirish Keskar and Tabarak Khan and Logan Kilpatrick and Jong Wook Kim and Christina Kim and Yongjik Kim and Hendrik Kirchner and Jamie Ryan Kiros and Matthew Knight and Daniel Kokotajlo and Lukasz Kondraciuk and Andrew Kondrich and Aris Konstantinidis and Kyle Kosic and Gretchen Krueger and Vishal Kuo and Michael Lampe and Ikai Lan and Teddy Lee and Jan Leike and Jade Leung and Daniel Levy and Chak Li and Rachel Lim and Molly Lin and Stephanie Lin and Ma-teusz Litwin and Theresa Lopez and Ryan Lowe and Patricia Lue and Anna Makanju and Kim Malfacini and Sam Manning and Todor Markov and Yaniv Markovski and Bianca Martin and Katie Mayer and Andrew Mayne and Bob McGrew and Scott Mayer McKinney and Christine McLeavey and Paul McMillan and Jake McNeil and David Medina and Aalok Mehta and Jacob Menick and Luke Metz and An-drey Mishchenko and Pamela Mishkin and Vinnie Monaco and Evan Morikawa and Daniel P. Mossing and Tong Mu and Mira Murati and Oleg Murk and David M'ely and Ashvin Nair and Reiichiro Nakano and Rajeev Nayak and Arvind Neelakantan and Richard Ngo and Hyeonwoo Noh and Ouyang Long and Cullen O'Keefe and Jakub W. Pachocki and Alex Paino and Joe Palermo and Ashley Pantuliano and Giambattista Parascandolo and Joel Parish and Emy Parparita and Alexandre Passos and Mikhail Pavlov and Andrew Peng and Adam Perelman and Filipe de Avila Belbute Peres and Michael Petrov and Henrique Pond{\'e} de Oliveira Pinto and Michael Pokorny and Michelle Pokrass and Vitchyr H. Pong and Tolly Powell and Alethea Power and Boris Power and Elizabeth Proehl and Raul Puri and Alec Radford and Jack W. Rae and Aditya Ramesh and Cameron Raymond and Francis Real and Kendra Rimbach and Carl Ross and Bob Rotsted and Henri Roussez and Nick Ryder and Mario D. Saltarelli and Ted Sanders and Shibani Santurkar and Girish Sastry and Heather Schmidt and David Schnurr and John Schulman and Daniel Selsam and Kyla Sheppard and Toki Sherbakov and Jessica Shieh and Sarah Shoker and Pranav Shyam and Szymon Sidor and Eric Sigler and Maddie Simens and Jordan Sitkin and Katarina Slama and Ian Sohl and Benjamin Sokolowsky and Yang Song and Natalie Staudacher and Felipe Petroski Such and Natalie Summers and Ilya Sutskever and Jie Tang and Nikolas A. Tezak and Madeleine Thompson and Phil Tillet and Amin Tootoonchian and Elizabeth Tseng and Preston Tuggle and Nick Turley and Jerry Tworek and Juan Felipe Cer'on Uribe and Andrea Vallone and Arun Vijayvergiya and Chelsea Voss and Carroll L. Wainwright and Justin Jay Wang and Alvin Wang and Ben Wang and Jonathan Ward and Jason Wei and CJ Weinmann and Akila Welihinda and Peter Welinder and Jiayi Weng and Lilian Weng and Matt Wiethoff and Dave Willner and Clemens Winter and Samuel Wolrich and Hannah Wong and Lauren Workman and Sherwin Wu and Jeff Wu and Michael Wu and Kai Xiao and Tao Xu and Sarah Yoo and Kevin Yu and Qim-ing Yuan and Wojciech Zaremba and Rowan Zellers and Chong Zhang and Marvin Zhang and Shengjia Zhao and Tianhao Zheng and Juntang Zhuang and William Zhuk and Barret Zoph},
  year={2023},
  url={https://api.semanticscholar.org/CorpusID:257532815}
}

@inproceedings{Ouyang2022TrainingLM,
author = {Ouyang, Long and Wu, Jeff and Jiang, Xu and Almeida, Diogo and Wainwright, Carroll L. and Mishkin, Pamela and Zhang, Chong and Agarwal, Sandhini and Slama, Katarina and Ray, Alex and Schulman, John and Hilton, Jacob and Kelton, Fraser and Miller, Luke and Simens, Maddie and Askell, Amanda and Welinder, Peter and Christiano, Paul and Leike, Jan and Lowe, Ryan},
title = {Training language models to follow instructions with human feedback},
year = {2022},
isbn = {9781713871088},
publisher = {Curran Associates Inc.},
address = {Red Hook, NY, USA},
booktitle = {Proceedings of the 36th International Conference on Neural Information Processing Systems},
articleno = {2011},
numpages = {15},
location = {New Orleans, LA, USA},
series = {NIPS '22},
url={https://api.semanticscholar.org/CorpusID:246426909}
}

@article{papafragou_scalar_2003,
	title = {Scalar implicatures: experiments at the semantics–pragmatics interface},
	volume = {86},
	issn = {0010-0277},
	url = {https://www.sciencedirect.com/science/article/pii/S0010027702001798},
	doi = {https://doi.org/10.1016/S0010-0277(02)00179-8},
	number = {3},
	journal = {Cognition},
	author = {Papafragou, Anna and Musolino, Julien},
	year = {2003},
	pages = {253--282},
}

@INPROCEEDINGS{park2018,
  author={Park, Dong Huk and Hendricks, Lisa Anne and Akata, Zeynep and Rohrbach, Anna and Schiele, Bernt and Darrell, Trevor and Rohrbach, Marcus},
  booktitle={2018 IEEE/CVF Conference on Computer Vision and Pattern Recognition}, 
  title={Multimodal Explanations: Justifying Decisions and Pointing to the Evidence}, 
  year={2018},
  volume={},
  number={},
  pages={8779-8788},
  doi={10.1109/CVPR.2018.00915}}

@inbook{inbook,
author = {Pearl, Judea},
year = {2021},
month = {12},
pages = {427-438},
title = {Causal and Counterfactual Inference},
isbn = {9780262366175},
doi = {10.7551/mitpress/11252.003.0044}
}

@inproceedings{Radford2019LanguageMA,
  title={Language Models are Unsupervised Multitask Learners},
  author={Alec Radford and Jeff Wu and Rewon Child and David Luan and Dario Amodei and Ilya Sutskever},
  year={2019},
  url={https://api.semanticscholar.org/CorpusID:160025533}
}

@inproceedings{ribeiro-etal-2016-trust,
    title = "``Why Should {I} Trust You?'': Explaining the Predictions of Any Classifier",
    author = "Ribeiro, Marco  and
      Singh, Sameer  and
      Guestrin, Carlos",
    editor = "DeNero, John  and
      Finlayson, Mark  and
      Reddy, Sravana",
    booktitle = "Proceedings of the 2016 Conference of the North {A}merican Chapter of the Association for Computational Linguistics: Demonstrations",
    month = jun,
    year = "2016",
    address = "San Diego, California",
    publisher = "Association for Computational Linguistics",
    url = "https://aclanthology.org/N16-3020/",
    doi = "10.18653/v1/N16-3020",
    pages = "97--101"
}

@inproceedings{ross-etal-2021-explaining,
    title = "Explaining {NLP} Models via Minimal Contrastive Editing ({M}i{CE})",
    author = "Ross, Alexis  and
      Marasovi{\'c}, Ana  and
      Peters, Matthew",
    editor = "Zong, Chengqing  and
      Xia, Fei  and
      Li, Wenjie  and
      Navigli, Roberto",
    booktitle = "Findings of the Association for Computational Linguistics: ACL-IJCNLP 2021",
    month = aug,
    year = "2021",
    address = "Online",
    publisher = "Association for Computational Linguistics",
    url = "https://aclanthology.org/2021.findings-acl.336/",
    doi = "10.18653/v1/2021.findings-acl.336",
    pages = "3840--3852"
}

@article{sauerland_scalar_2004,
	title = {Scalar {Implicatures} in {Complex} {Sentences}},
	volume = {27},
	issn = {1573-0549},
	url = {https://doi.org/10.1023/B:LING.0000023378.71748.db},
	doi = {10.1023/B:LING.0000023378.71748.db},
	number = {3},
	journal = {Linguistics and Philosophy},
	author = {Sauerland, Uli},
	month = jun,
	year = {2004},
	pages = {367--391},
}

@article{Schoenegger2024AnEO,
  author       = {Loris Schoenegger and
                  Yuxi Xia and
                  Benjamin Roth},
  title        = {An Evaluation of Explanation Methods for Black-Box Detectors of Machine-Generated
                  Text},
  journal      = {CoRR},
  volume       = {abs/2408.14252},
  year         = {2024},
  url          = {https://doi.org/10.48550/arXiv.2408.14252},
  doi          = {10.48550/ARXIV.2408.14252},
  eprinttype    = {arXiv},
  eprint       = {2408.14252},
  bibsource    = {dblp computer science bibliography, https://dblp.org}
}

@inproceedings{treviso-etal-2023-crest,
    title = "{CREST}: A Joint Framework for Rationalization and Counterfactual Text Generation",
    author = "Treviso, Marcos  and
      Ross, Alexis  and
      Guerreiro, Nuno M.  and
      Martins, Andr{\'e}",
    editor = "Rogers, Anna  and
      Boyd-Graber, Jordan  and
      Okazaki, Naoaki",
    booktitle = "Proceedings of the 61st Annual Meeting of the Association for Computational Linguistics (Volume 1: Long Papers)",
    month = jul,
    year = "2023",
    address = "Toronto, Canada",
    publisher = "Association for Computational Linguistics",
    url = "https://aclanthology.org/2023.acl-long.842/",
    doi = "10.18653/v1/2023.acl-long.842",
    pages = "15109--15126"
}

@inproceedings{wang-etal-2024-coxql,
    title = "{C}o{XQL}: A Dataset for Parsing Explanation Requests in Conversational {XAI} Systems",
    author = {Wang, Qianli  and
      Anikina, Tatiana  and
      Feldhus, Nils  and
      Ostermann, Simon  and
      M{\"o}ller, Sebastian},
    editor = "Al-Onaizan, Yaser  and
      Bansal, Mohit  and
      Chen, Yun-Nung",
    booktitle = "Findings of the Association for Computational Linguistics: EMNLP 2024",
    month = nov,
    year = "2024",
    address = "Miami, Florida, USA",
    publisher = "Association for Computational Linguistics",
    url = "https://aclanthology.org/2024.findings-emnlp.76/",
    doi = "10.18653/v1/2024.findings-emnlp.76",
    pages = "1410--1422"
}

@inproceedings{wang-etal-2025-fitcf,
    title = "{F}it{CF}: A Framework for Automatic Feature Importance-guided Counterfactual Example Generation",
    author = {Wang, Qianli  and
      Feldhus, Nils  and
      Ostermann, Simon  and
      Villa-Arenas, Luis Felipe  and
      M{\"o}ller, Sebastian  and
      Schmitt, Vera},
    editor = "Che, Wanxiang  and
      Nabende, Joyce  and
      Shutova, Ekaterina  and
      Pilehvar, Mohammad Taher",
    booktitle = "Findings of the Association for Computational Linguistics: ACL 2025",
    month = jul,
    year = "2025",
    address = "Vienna, Austria",
    publisher = "Association for Computational Linguistics",
    url = "https://aclanthology.org/2025.findings-acl.64/",
    doi = "10.18653/v1/2025.findings-acl.64",
    pages = "1176--1191",
    ISBN = "979-8-89176-256-5"
}

@inproceedings{wu-etal-2021-polyjuice,
    title = "Polyjuice: Generating Counterfactuals for Explaining, Evaluating, and Improving Models",
    author = "Wu, Tongshuang  and
      Ribeiro, Marco Tulio  and
      Heer, Jeffrey  and
      Weld, Daniel",
    editor = "Zong, Chengqing  and
      Xia, Fei  and
      Li, Wenjie  and
      Navigli, Roberto",
    booktitle = "Proceedings of the 59th Annual Meeting of the Association for Computational Linguistics and the 11th International Joint Conference on Natural Language Processing (Volume 1: Long Papers)",
    month = aug,
    year = "2021",
    address = "Online",
    publisher = "Association for Computational Linguistics",
    url = "https://aclanthology.org/2021.acl-long.523/",
    doi = "10.18653/v1/2021.acl-long.523",
    pages = "6707--6723"
}

@inproceedings{zheng_judge,
author = {Zheng, Lianmin and Chiang, Wei-Lin and Sheng, Ying and Zhuang, Siyuan and Wu, Zhanghao and Zhuang, Yonghao and Lin, Zi and Li, Zhuohan and Li, Dacheng and Xing, Eric P. and Zhang, Hao and Gonzalez, Joseph E. and Stoica, Ion},
title = {Judging LLM-as-a-judge with MT-bench and Chatbot Arena},
year = {2023},
publisher = {Curran Associates Inc.},
address = {Red Hook, NY, USA},
booktitle = {Proceedings of the 37th International Conference on Neural Information Processing Systems},
articleno = {2020},
numpages = {29},
location = {New Orleans, LA, USA},
series = {NIPS '23},
url = {https://dl.acm.org/doi/10.5555/3666122.3668142}
}

@article{zuo2025evadellmbasedexplanationgeneration,
  author       = {Longfei Zuo and
                  Barbara Plank and
                  Siyao Peng},
  title        = {{EVADE:} LLM-Based Explanation Generation and Validation for Error
                  Detection in {NLI}},
  journal      = {CoRR},
  volume       = {abs/2511.08949},
  year         = {2025},
  url          = {https://doi.org/10.48550/arXiv.2511.08949},
  doi          = {10.48550/ARXIV.2511.08949},
  eprinttype    = {arXiv},
  eprint       = {2511.08949},
  bibsource    = {dblp computer science bibliography, https://dblp.org}
}
\appendix

\section{LLM-prompts for automatic counterfactual questions generation}\label{sec:llm-question-generation}

This appendix shows the prompt used to generate follow-up counterfactual yes/no questions on the StrategyQA dataset. We use exactly the same natural-language prompt for all models and both backends (OpenAI Chat Completions and local Hugging Face models). For OpenAI models, the prompt is passed as \texttt{system}/\texttt{user} messages, and for Hugging Face models we serialize the same messages via the tokenizer's chat template using the code in the main text.

\begin{tcolorbox}[mygraybox,title={System Instructions}] 
\small
Human: In the questions below, you will be asked to read a starter yes or no question and a robot's answer to the starter question. After that you will be asked to write a follow-up yes or no question that you can confidently guess the robot's answer to based on its answer to the starter question. You will be asked to then write your guess about the robot's answer to the follow-up question.

\medskip
Assistant: here is my response. okay.
\end{tcolorbox}

\begin{tcolorbox}[mygraybox,title={Template}]
\small
Human: Starter Question: \texttt{\{orig\_qn\}}\\
Robot's Answer to the Starter Question: \texttt{\{orig\_yn\}}\\
Follow-up Question:
\end{tcolorbox}

\begin{tcolorbox}[mygraybox,title={Format Requirement}]
\small
When you answer, output exactly two lines:\\
\texttt{Follow-up Question: <a yes/no question>}\\
\texttt{Guessed Answer: yes or no}\\
Do not include anything else.
\end{tcolorbox}

\section{LLM-prompts for presupposition flip counterfactual questions generation}\label{sec:presupp-prompt}
\begin{tcolorbox}[mygraybox,title={System Instructions}]
\small
You are a precise editor that rewrites yes/no questions into counterfactual
questions with if-clauses while flipping the original answer. Follow the user's
format, be concise, and do not add extra commentary or bullets.
\end{tcolorbox}

\begin{tcolorbox}[mygraybox,title={User Instructions}]
\small
\textbf{Prompt/ instructions for LLM:}

Step 1: Read the original yes/no question and the answer.\\
Step 2: Flip the original answer.\\
Step 3: Work out how to make the answer true.\\
Step 4: Write a new question that asks if the answer would be true if the action you worked out in Step 3 was performed. Change the original questions to new questions of unreal conditions with counterfactual presuppositions, using if clauses. Do not change the meaning of questions in new questions.\\
Step 5: Write the new answer to the new questions.

\vspace{0.5em}
Answer each initial question with the following format:

\texttt{Original question: <original question>}\\
\texttt{Original answer: <original answer>}\\
\texttt{Step1: The new answer should be <yes/no>, so \dots}\\
\texttt{Step2: How to make \dots: \dots}\\
\texttt{New question: <new question>}

\vspace{0.5em}
Here are some examples:

Original question: Would the top of Mount Fuji stick out of the Sea of Japan?\\
Original answer: Yes\\
New question: If global sea level were 4,000 meters higher, would the top of Mount Fuji stick out of the Sea of Japan?\\
New answer: No

\vspace{0.5em}
Original question: Would an uninsured person be more likely than an insured person to decline a CT scan?\\
Original answer: Yes\\
New question: If CT scans were free at the point of care for everyone, would an uninsured person be more likely than an insured person to decline a CT scan?\\
New answer: No

\end{tcolorbox}

\section{Scalar scales} \label{sec:scales}

Table~\ref{tab:scalar-scales} lists the manually specified scalar scales used in our scalar adjustment operations. The scales are defined on the basis of scalar inventories discussed in previous work on scalar expressions and implicatures \citep{carston98,papafragou_scalar_2003,sauerland_scalar_2004}.

\begin{table*}[t]
\centering
\footnotesize
\begin{tabularx}{\textwidth}{lX}
\toprule
\textbf{Scale} & \textbf{Ordered items (stronger $\rightarrow$ weaker)} \\
\midrule
Quantifiers &
\emph{all}, \emph{almost all}, \emph{most}, \emph{a majority of}, \emph{many}, \emph{several}, \emph{some}, \emph{a few}, \emph{few}, \emph{hardly any} \\
\addlinespace
Universal quantifiers &
\emph{every}, \emph{each} \\
\addlinespace
Existential quantifiers&
\emph{any}, \emph{at least one}, \emph{at least some} \\
\addlinespace
Frequencies &
\emph{always}, \emph{almost always}, \emph{usually}, \emph{often}, \emph{frequently}, \emph{sometimes}, \emph{occasionally}, \emph{rarely}, \emph{seldom}, \emph{hardly ever}, \emph{never} \\
\addlinespace
Epistemic modality &
\emph{certainly}, \emph{definitely}, \emph{undoubtedly}, \emph{almost certainly}, \emph{very likely}, \emph{likely}, \emph{probably}, \emph{plausibly}, \emph{possibly}, \emph{maybe}, \emph{perhaps} \\
\addlinespace
Deontic modality &
\emph{must}, \emph{have to}, \emph{be required to}, \emph{should}, \emph{ought to}, \emph{be supposed to}, \emph{be recommended to}, \emph{may}, \emph{can} \\
\addlinespace
Necessity / possibility &
\emph{necessary}, \emph{required}, \emph{essential}, \emph{recommended}, \emph{advisable}, \emph{optional}, \emph{possible} \\
\addlinespace
Attitude / success &
\emph{succeed in}, \emph{manage to}, \emph{be able to}, \emph{try to}, \emph{attempt to}, \emph{want to} \\
\addlinespace
Evidential / reporting &
\emph{prove that}, \emph{demonstrate that}, \emph{show that}, \emph{find that}, \emph{report that}, \emph{suggest that}, \emph{speculate that} \\
\addlinespace
Claim strength / hedging &
\emph{it is certain that}, \emph{it is clear that}, \emph{it is likely that}, \emph{it is probable that}, \emph{it is plausible that}, \emph{it is possible that} \\
\addlinespace
Degree adverbs &
\emph{completely}, \emph{totally}, \emph{entirely}, \emph{highly}, \emph{very}, \emph{quite}, \emph{fairly}, \emph{rather}, \emph{somewhat}, \emph{slightly} \\
\addlinespace
Evaluative adjectives &
\emph{perfect}, \emph{excellent}, \emph{great}, \emph{good}, \emph{acceptable}, \emph{adequate}, \emph{poor} \\
\addlinespace
Temporal immediacy &
\emph{immediately}, \emph{right away}, \emph{soon}, \emph{eventually}, \emph{someday} \\
\addlinespace
Inclusion &
\emph{including all}, \emph{including most}, \emph{including many}, \emph{including some} \\
\bottomrule
\end{tabularx}
\caption{Manually specified scalar scales used for scalar adjustment. Within each row, items are ordered from stronger to weaker.}
\label{tab:scalar-scales}
\end{table*}

\section{LLM-prompts for explanations generation}\label{sec:explanation-generator}

This appendix shows the prompt used for generating chain-of-thought and post-hoc explanations on the StrategyQA dataset. We use exactly the same natural-language prompt for all three models and both backends. For OpenAI models, the prompt is passed as \texttt{system}/\texttt{user} chat messages, while for Hugging Face models we serialize the same messages using the tokenizer's chat template.

\begin{tcolorbox}[mygraybox,title={Prompt: Explanation Generator}]
\small
Generate explanations (CoT / Post-hoc).

\medskip
\textbf{System instruction for CoT (\texttt{COT\_SYSTEM}):}

Human: In the questions below, you will be asked to first think step by step and generate a trace of reasoning, and then end with your final answer exactly with ``So the answer is ...'' yes or no. Let's think step by step. Strictly follow the example format below and do not say anything else.

\medskip
Assistant: here is my response. okay.

\medskip
\textbf{System instruction for post-hoc explanations (\texttt{POSTHOC\_SYSTEM}):}

Human: In the questions below, you will be asked to first answer the question (yes or no) and then generate a justification for your answer. Your justification should end with "So the answer is ..."\ yes or no.

\medskip
Assistant: here is my response. okay.

\medskip
\textbf{User message template for CoT:}

Human: Q: Yes or no: \texttt{\{question\}}\\
A: \\
Assistant: here is my response.

\medskip
\textbf{User message template for post-hoc explanations:}

Human: Q: Yes or no: \texttt{\{question\}}\\
A: \\
Assistant: here is my response.
\end{tcolorbox}

\section{User study guideline}\label{sec:user-guideline}
We recruited 28 volunteer participants (B.Sc., M.Sc., and PhD students) with English reading proficiency at CEFR level C1 or above. Before the study, participants reported (i) their familiarity with LLMs on a 5-point Likert scale and (ii) whether they were familiar with explainable AI (XAI). Most participants frequently used LLMs in their work: 18/28 selected 5 (“very familiar / frequent use”), 4/28 selected 4 (“regular use in daily life”), and the remaining 6/28 selected 3 (“occasional use”). In contrast, only 2/28 reported in-depth knowledge of XAI; the rest indicated limited familiarity or only a basic awareness. Participation was fully voluntary, and participants were generally interested in NLP and AI research. The experiment was conducted online via a purpose-built web service.

\begin{figure*}[t]
  \centering
  \includegraphics[width=0.9\textwidth]{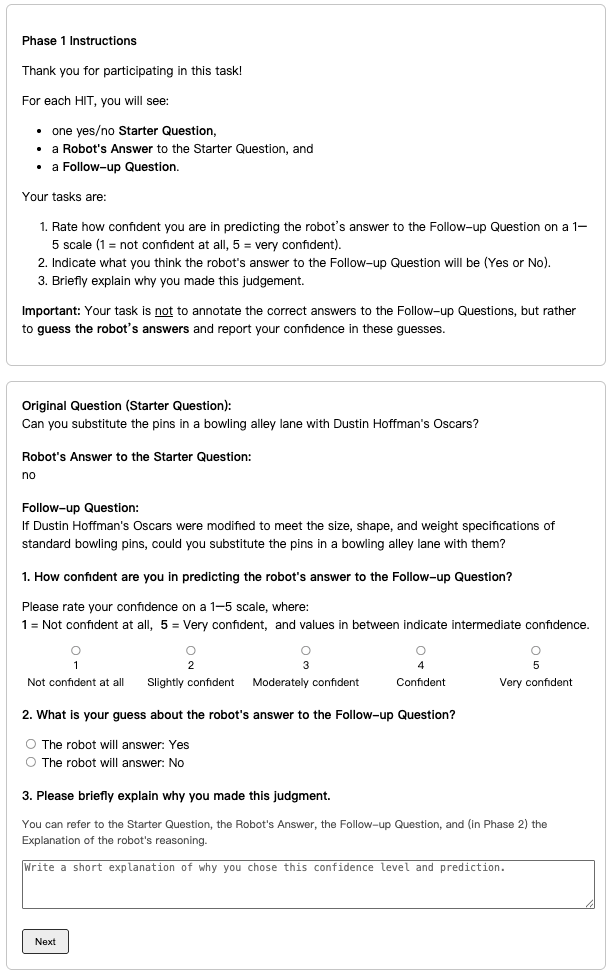}
  \caption{User study interface: instruction for phase 1.}
  \label{fig:phase1-intro}
\end{figure*}

\begin{figure*}[t]
  \centering
  \includegraphics[width=0.75\textwidth]{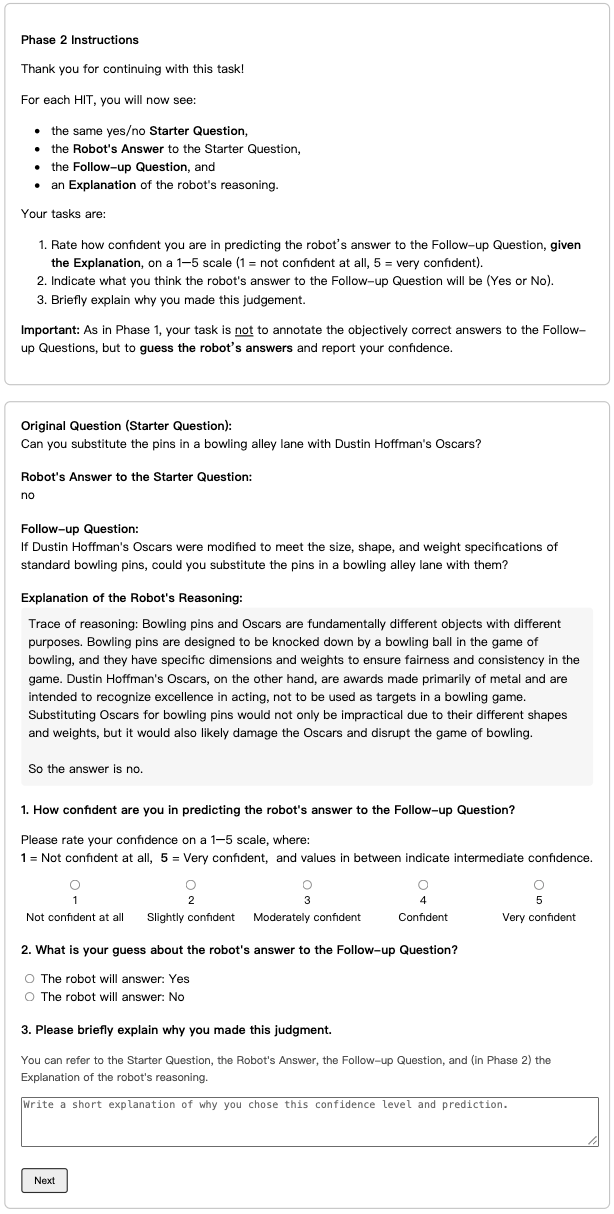}
  \caption{User study interface: instruction for phase 2.}
  \label{fig:phase2-intro}
\end{figure*}

In the following, we provide the full annotation guideline used in our user study. Before starting the study, each participant is shown this guideline, which includes a general introduction to the task, detailed instructions, and illustrative examples. Participants are encouraged to ask questions both before and during the study. Screenshots of the user study interface are shown in Figures~\ref{fig:phase1-intro} and~\ref{fig:phase2-intro}.

\subsection{Overview of the task}

In this study, you will see short question--answer pairs involving a ``robot'' (a large language model). Your job is to reason about what the robot will answer to a follow-up yes/no question. There are two phases:
\paragraph{Phase 1}: You see a \textit{Starter Question}, the \textit{Robot's Answer}, and a \textit{Follow-up Question}.
\paragraph{Phase 2}: You see the same elements plus an \textit{Explanation} of the robot's reasoning for the follow-up question.

In every trial, you will:
\begin{itemize}
    \item rate how confident you are that you can reason about the robot's answer to the follow-up question;
    \item guess the robot's answer to the follow-up question (Yes or No);
    \item provide a short free-text explanation of how you made your guess.
\end{itemize}

\subsection{\textit{Very important:} Guess the robot's answer, not your own opinion}

Your task is \textbf{not} to say what you think is objectively correct. Your task is to guess what the robot will answer to the follow-up question. Even if you believe the robot is wrong or unreasonable, you should still try to infer its likely answer. Base your guess on: the \textit{Starter Question}, the \textit{Robot's Answer}, and (in Phase 2) the \textit{Explanation}.

In your explanations, please focus on how you inferred the robot's behavior, not on what you personally think is true about the world.

\subsection{Confidence rating and forced guess}

For each trial, you will first answer:

\begin{quote}
\textit{``How confident are you that you can reason about what the robot will answer to the follow-up question?''}
\end{quote}

You will choose a number on a 1--5 scale (e.g., 1 = Not confident at all, 5 = Extremely confident).

After choosing a confidence level, you must still select one of:
\begin{itemize}
    \item The robot will answer: Yes
    \item The robot will answer: No
\end{itemize}

There are no ``I don't know'' or ``cannot tell'' options in the multiple-choice part.

If you feel that you really cannot predict the robot's answer at all:
\begin{itemize}
    \item choose the lowest confidence level;
    \item still pick Yes or No as a forced guess;
    \item in the free-text explanation, clearly say that you cannot guess, e.g.: \textcolor{MidnightBlue}{\textit{``I can't guess the robot's answer; I chose `Yes' randomly.''}}
\end{itemize}

This helps us distinguish between informed reasoning and random guessing.

\subsection{What to write in your free-text explanation}

After you have given your confidence rating and your Yes/No guess, you will write a short explanation (typically 1--3 sentences). Here are some general tips:

\begin{itemize}
    \item Keep it simple and concise, but concrete.
    \item Explain why you think the robot will answer Yes or No.
    \item Focus on how you used the robot's previous answer and (in Phase 2) the explanation.
    \item You do not need to write a long essay; 1--3 sentences are enough.
\end{itemize}

\subsection{Phase 1 guidelines (without explanation)}

Typical patterns you may use (and can mention in your explanation):

\paragraph{Same conditions $\rightarrow$ same answer:}
\textcolor{MidnightBlue}{\textit{``The follow-up question is very similar to the starter question, so I think the robot will give the same answer.''}}

\paragraph{Opposite conditions $\rightarrow$ opposite answer:}
\textcolor{MidnightBlue}{\textit{``The follow-up question reverses the condition in the starter question, so I expect the robot to give the opposite answer.''}}

\paragraph{Inference from the Robot's Answer:}
\textcolor{MidnightBlue}{\textit{``From the robot's answer `no' to the starter question, I infer that it thinks X is unlikely, so I expect it will also answer `no' to the follow-up.''}}

\paragraph{Unclear / low confidence:}
\textcolor{MidnightBlue}{\textit{``I'm not confident. The robot's answer to the starter question does not clearly tell me how it will answer the follow-up, so this is mostly a guess.''}}

\subsection{Phase 2 guidelines (with explanation)}

For your free-text explanation in Phase 2 , you are encouraged to refer directly to the explanation. You can quote short parts in quotation marks, for example:
\textcolor{MidnightBlue}{
\textit{``The explanation says `Rede Globo is a Brazilian television network', so I expect the robot to answer `Yes' when asked if the anchors speak Portuguese.''}}

You can combine the explanation with your own reasoning:
\textcolor{MidnightBlue}{
\textit{``The explanation says `X is the official language of Y', which suggests the robot believes people in Y usually speak X. Therefore, I think it will answer `Yes' to the follow-up question.''}}

\subsection{Examples of good and less useful explanations}

\textbf{Good explanations} refer to the robot's answer and/or the explanation and make it clear how you predicted the robot's behavior. Here are some examples of good free-text explanations:

Example 1:
\textcolor{MidnightBlue}{
\textit{``From the robot's `no' to the starter question and the explanation that Chinese is not commonly spoken in Brazil, I think it will answer `yes' when asked about Portuguese.''}}

Example 2:
\textcolor{MidnightBlue}{\textit{``The explanation states `X is unlikely', so I expect the robot to keep the same negative answer for the follow-up.''}}

Try to avoid explanations that rely only on personal opinions without referring to the robot, that simply repeat the question, or that say nothing about your reasoning. If it really is just a guess, please still explain why you cannot reason about the robot's answer and mark low confidence. Here are some examples of less informative explanations:

Example 1:
\textcolor{MidnightBlue}{\textit{``I think the correct answer is yes.''}}

Example 2:
\textcolor{MidnightBlue}{\textit{``Because of the question.''}}

Example 3:
\textcolor{MidnightBlue}{\textit{``Just a guess.''}}

\subsection{Summary}

\begin{itemize}
    \item Always guess the robot's answer, not your own belief about what is true.
    \item Always provide a Yes/No guess, even if you are not confident.
    \item If you really cannot predict the answer:
    \begin{itemize}
        \item choose ``not confident'';
        \item still choose Yes or No; and
        \item write in your explanation that you cannot guess and why.
    \end{itemize}
    \item In Phase 1, base your reasoning mainly on the \textit{Starter Question} and the \textit{Robot's Answer}.
    \item In Phase 2, also use the \textit{Explanation}, and feel free to quote it in quotation marks.
    \item Keep explanations short but specific, focusing on how you infer the robot's answer.
\end{itemize}

\section{Full results for LLM-as-a-judge counterfactual simulation}\label{sec:full-judge-results}

This appendix reports the full per-condition results for our LLM-as-a-judge counterfactual simulation. 

We report three complementary accuracy-based metrics: (i) \textbf{guess\_rate}, the proportion of items judged as \texttt{can\_guess}; (ii) \textbf{selective\_accuracy}, accuracy conditioned on guessing; and (iii) \textbf{overall accuracy}, accuracy on all items, counting \texttt{cannot\_guess} as incorrect. Results using two LLM judges, \texttt{gpt-3.5-turbo} and \texttt{gpt-4.1}, are reported in Table~\ref{tab:cf-sim-gpt35} and Table~\ref{tab:cf-sim-gpt41}.

\begin{table*}[t]
\centering
\resizebox{\textwidth}{!}{
\begin{tabular}{l l l l l
             r r r r r r}
\toprule
\textbf{expl} &
\textbf{cfq\_type} &
\textbf{P1\_guess\_rate} &
\textbf{P1\_acc} &
\textbf{P1\_sel\_acc} &
\textbf{P2\_guess\_rate} &
\textbf{P2\_acc} &
\textbf{P2\_sel\_acc} &
\textbf{$\Delta$ guess\_rate} &
\textbf{$\Delta$ acc} &
\textbf{$\Delta$ sel\_acc} \\
\midrule
cot (gpt-3.5) & gpt-3.5 & 0.505 & 0.368 & 0.000 & 0.901 & 0.624 & 0.737 & 0.396 & \cellcolor{blue!15}0.256 & 0.008 \\
cot (gpt-3.5) & gpt-4 & 0.623 & 0.479 & 0.769 & 0.940 & 0.712 & 0.757 & 0.317 & \cellcolor{blue!30}0.233 & -0.012 \\
cot (gpt-3.5) & Llama-3.3 & 0.562 & 0.414 & 0.735 & 0.934 & 0.684 & 0.732 & 0.371 & \cellcolor{blue!5}0.270 & -0.003 \\
cot (gpt-3.5) & presupp\_flip & 0.434 & 0.267 & 0.614 & 0.894 & 0.562 & 0.628 & 0.460 & \cellcolor{pink!60}0.295 & 0.014 \\
cot (gpt-3.5) & contextual & 0.479 & 0.315 & 0.658 & 0.832 & 0.587 & 0.706 & 0.353 & \cellcolor{pink!15}0.272 & 0.048 \\
cot (gpt-3.5) & lexical & 0.244 & 0.134 & 0.550 & 0.820 & 0.507 & 0.617 & 0.576 & \cellcolor{pink!95}0.442 & 0.067 \\
cot (gpt-3.5) & scalar & 0.514 & 0.304 & 0.590 & 0.869 & 0.590 & 0.679 & 0.355 & \cellcolor{pink!35}0.286 & 0.089 \\
\midrule
posthoc (gpt-3.5) & gpt-3.5 & 0.485 & 0.358 & 0.738 & 0.842 & 0.626 & 0.743 & 0.357 & \cellcolor{blue!5}0.268 & 0.005 \\
posthoc (gpt-3.5) & gpt-4 & 0.552 & 0.387 & 0.702 & 0.862 & 0.647 & 0.750 & 0.310 & \cellcolor{blue!15}0.260 & 0.048 \\
posthoc (gpt-3.5) & Llama-3.3 & 0.517 & 0.374 & 0.723 & 0.797 & 0.613 & 0.768 & 0.280 & \cellcolor{blue!30}0.239 & 0.045 \\
posthoc (gpt-3.5) & presupp\_flip & 0.446 & 0.311 & 0.698 & 0.810 & 0.663 & 0.652 & 0.364 & \cellcolor{pink!60}0.352 & -0.046 \\
posthoc (gpt-3.5) & contextual & 0.526 & 0.339 & 0.643 & 0.901 & 0.643 & 0.714 & 0.375 & \cellcolor{pink!35}0.304 & 0.071 \\
posthoc (gpt-3.5) & lexical & 0.448 & 0.339 & 0.758 & 0.892 & 0.766 & 0.743 & 0.444 & \cellcolor{pink!95}0.427 & -0.015 \\
posthoc (gpt-3.5) & scalar & 0.628 & 0.508 & 0.809 & 0.861 & 0.801 & 0.790 & 0.233 & \cellcolor{pink!15}0.293 & -0.019 \\
\midrule
cot (gpt-4) & gpt-3.5 & 0.442 & 0.358 & 0.808 & 0.772 & 0.659 & 0.854 & 0.330 & \cellcolor{blue!5}0.301 & 0.046 \\
cot (gpt-4) & gpt-4 & 0.566 & 0.476 & 0.843 & 0.914 & 0.807 & 0.882 & 0.348 & \cellcolor{pink!60}0.331 & 0.040 \\
cot (gpt-4) & Llama-3.3 & 0.550 & 0.463 & 0.842 & 0.879 & 0.763 & 0.868 & 0.329 & \cellcolor{blue!15}0.300 & 0.026 \\
cot (gpt-4) & presupp\_flip & 0.406 & 0.338 & 0.833 & 0.764 & 0.645 & 0.844 & 0.358 & \cellcolor{pink!15}0.307 & 0.011 \\
cot (gpt-4) & contextual & 0.674 & 0.603 & 0.894 & 0.948 & 0.818 & 0.863 & 0.273 & \cellcolor{blue!30}0.215 & -0.031 \\
cot (gpt-4) & lexical & 0.466 & 0.380 & 0.815 & 0.804 & 0.712 & 0.818 & 0.338 & \cellcolor{pink!95}0.332 & 0.003 \\
cot (gpt-4) & scalar & 0.543 & 0.458 & 0.868 & 0.880 & 0.769 & 0.858 & 0.337 & \cellcolor{pink!35}0.311 & -0.010 \\
\midrule
posthoc (gpt-4) & gpt-3.5 & 0.456 & 0.364 & 0.799 & 0.768 & 0.660 & 0.860 & 0.312 & \cellcolor{pink!15}0.296 & 0.061 \\
posthoc (gpt-4) & gpt-4 & 0.590 & 0.483 & 0.818 & 0.906 & 0.801 & 0.884 & 0.316 & \cellcolor{pink!35}0.318 & 0.066 \\
posthoc (gpt-4) & Llama-3.3 & 0.543 & 0.343 & 0.873 & 0.838 & 0.732 & 0.882 & 0.295 & \cellcolor{pink!90}0.389 & 0.009 \\
posthoc (gpt-4) & presupp\_flip & 0.553 & 0.432 & 0.781 & 0.831 & 0.717 & 0.845 & 0.278 & \cellcolor{blue!5}0.285 & 0.064 \\
posthoc (gpt-4) & contextual & 0.671 & 0.588 & 0.877 & 0.948 & 0.805 & 0.849 & 0.277 & \cellcolor{blue!30}0.217 & -0.028 \\
posthoc (gpt-4) & lexical & 0.332 & 0.358 & 0.830 & 0.698 & 0.727 & 0.856 & 0.366 & \cellcolor{pink!60}0.369 & 0.026 \\
posthoc (gpt-4) & scalar & 0.658 & 0.546 & 0.830 & 0.820 & 0.792 & 0.846 & 0.162 & \cellcolor{blue!15}0.246 & 0.016 \\
\midrule
cot (Llama-3.3) & gpt-3.5 & 0.478 & 0.395 & 0.826 & 0.768 & 0.673 & 0.877 & 0.290 & \cellcolor{blue!30}0.278 & 0.051 \\
cot (Llama-3.3) & gpt-4 & 0.607 & 0.511 & 0.842 & 0.908 & 0.799 & 0.880 & 0.301 & \cellcolor{blue!15}0.288 & 0.038 \\
cot (Llama-3.3) & Llama-3.3 & 0.523 & 0.445 & 0.851 & 0.897 & 0.774 & 0.864 & 0.374 & \cellcolor{pink!35}0.329 & 0.013 \\
cot (Llama-3.3) & presupp\_flip & 0.365 & 0.312 & 0.854 & 0.762 & 0.655 & 0.860 & 0.397 & \cellcolor{pink!60}0.343 & 0.006 \\
cot (Llama-3.3) & contextual & 0.763 & 0.349 & 0.457 & 0.942 & 0.628 & 0.562 & 0.179 & \cellcolor{blue!5}0.289 & 0.105 \\
cot (Llama-3.3) & lexical & 0.434 & 0.334 & 0.770 & 0.793 & 0.688 & 0.889 & 0.359 & \cellcolor{pink!90}0.354 & 0.119 \\
cot (Llama-3.3) & scalar & 0.593 & 0.403 & 0.848 & 0.765 & 0.723 & 0.815 & 0.172 & \cellcolor{pink!15}0.320 & -0.033 \\
\midrule
posthoc (Llama-3.3) & gpt-3.5 & 0.463 & 0.361 & 0.780 & 0.773 & 0.662 & 0.856 & 0.310 & \cellcolor{blue!15}0.301 & 0.076 \\
posthoc (Llama-3.3) & gpt-4 & 0.604 & 0.491 & 0.813 & 0.904 & 0.791 & 0.875 & 0.300 & \cellcolor{blue!30}0.300 & 0.062 \\
posthoc (Llama-3.3) & Llama-3.3 & 0.529 & 0.424 & 0.801 & 0.899 & 0.778 & 0.865 & 0.379 & \cellcolor{pink!35}0.354 & 0.064 \\
posthoc (Llama-3.3) & presupp\_flip & 0.416 & 0.356 & 0.856 & 0.719 & 0.721 & 0.862 & 0.303 & \cellcolor{pink!90}0.365 & 0.006 \\
posthoc (Llama-3.3) & contextual & 0.741 & 0.237 & 0.319 & 0.942 & 0.600 & 0.318 & 0.201 & \cellcolor{pink!60}0.363 & -0.001 \\
posthoc (Llama-3.3) & lexical & 0.434 & 0.355 & 0.817 & 0.622 & 0.694 & 0.836 & 0.188 & \cellcolor{pink!15}0.339 & 0.019 \\
posthoc (Llama-3.3) & scalar & 0.653 & 0.462 & 0.861 & 0.821 & 0.788 & 0.838 & 0.168 & \cellcolor{blue!5}0.326 & -0.023 \\
\bottomrule
\end{tabular}
}
\caption{Per-condition results for counterfactual simulatability \textit{(judge: \texttt{gpt-3.5-turbo})}. Improvements are Phase~2 minus Phase~1.}
\label{tab:cf-sim-gpt35}
\end{table*}

\begin{table*}[t]
\centering
\resizebox{\textwidth}{!}{
\begin{tabular}{l l l l l
             r r r r r r}
\toprule
\textbf{expl} &
\textbf{cfq\_type} &
\textbf{P1\_guess\_rate} &
\textbf{P1\_acc} &
\textbf{P1\_sel\_acc} &
\textbf{P2\_guess\_rate} &
\textbf{P2\_acc} &
\textbf{P2\_sel\_acc} &
\textbf{$\Delta$ guess\_rate} &
\textbf{$\Delta$ acc} &
\textbf{$\Delta$ sel\_acc} \\
\midrule
cot (gpt-3.5) & gpt-3.5 & 0.607 & 0.445 & 0.733 & 0.764 & 0.586 & 0.768 & 0.157 & \cellcolor{pink!15}0.141 & 0.035 \\
cot (gpt-3.5) & gpt-4 & 0.740 & 0.541 & 0.731 & 0.859 & 0.650 & 0.757 & 0.119 & \cellcolor{blue!15}0.109 & 0.026 \\
cot (gpt-3.5) & Llama-3.3 & 0.706 & 0.528 & 0.748 & 0.841 & 0.639 & 0.760 & 0.135 & \cellcolor{blue!5}0.111 & 0.012 \\
cot (gpt-3.5) & PresupFlip & 0.664 & 0.401 & 0.604 & 0.854 & 0.571 & 0.587 & 0.190 & \cellcolor{pink!60}0.170 & -0.017 \\
cot (gpt-3.5) & Contextual & 0.941 & 0.718 & 0.763 & 0.949 & 0.730 & 0.769 & 0.008 & \cellcolor{blue!30}0.012 & 0.006 \\
cot (gpt-3.5) & Lexical & 0.418 & 0.301 & 0.720 & 0.586 & 0.462 & 0.766 & 0.168 & \cellcolor{pink!35}0.161 & 0.046 \\
cot (gpt-3.5) & Scalar & 0.670 & 0.516 & 0.770 & 0.788 & 0.719 & 0.785 & 0.118 & \cellcolor{pink!90}0.203 & 0.015 \\
\midrule
posthoc (gpt-3.5) & gpt-3 & 0.689 & 0.479 & 0.752 & 0.774 & 0.589 & 0.761 & 0.085 & \cellcolor{blue!15}0.110 & 0.009 \\
posthoc (gpt-3.5) & gpt-4 & 0.732 & 0.506 & 0.691 & 0.847 & 0.635 & 0.750 & 0.115 & \cellcolor{blue!5}0.129 & 0.059 \\
posthoc (gpt-3.5) & Llama-3.3 & 0.695 & 0.499 & 0.718 & 0.838 & 0.635 & 0.758 & 0.143 & \cellcolor{pink!15}0.136 & 0.040 \\
posthoc (gpt-3.5) & PresupFlip & 0.694 & 0.423 & 0.610 & 0.856 & 0.615 & 0.617 & 0.162 & \cellcolor{pink!90}0.192 & 0.007 \\
posthoc (gpt-3.5) & Contextual & 0.947 & 0.676 & 0.714 & 0.949 & 0.690 & 0.708 & 0.002 & \cellcolor{blue!30}0.014 & -0.006 \\
posthoc (gpt-3.5) & Lexical & 0.425 & 0.300 & 0.705 & 0.595 & 0.478 & 0.743 & 0.179 & \cellcolor{pink!60}0.178 & 0.038 \\
posthoc (gpt-3.5) & Scalar & 0.652 & 0.484 & 0.743 & 0.795 & 0.623 & 0.771 & 0.143 & \cellcolor{pink!35}0.139 & 0.028 \\
\midrule
cot (gpt-4) & gpt-3.5 & 0.591 & 0.488 & 0.826 & 0.799 & 0.693 & 0.868 & 0.208 & \cellcolor{pink!60}0.205 & 0.042 \\
cot (gpt-4) & gpt-4 & 0.768 & 0.660 & 0.860 & 0.919 & 0.809 & 0.881 & 0.151 & \cellcolor{blue!15}0.149 & 0.021 \\
cot (gpt-4) & Llama-3.3 & 0.707 & 0.602 & 0.851 & 0.896 & 0.782 & 0.873 & 0.189 & \cellcolor{pink!15}0.180 & 0.022 \\
cot (gpt-4) & PresupFlip & 0.638 & 0.530 & 0.831 & 0.880 & 0.716 & 0.814 & 0.242 & \cellcolor{pink!35}0.186 & -0.017 \\
cot (gpt-4) & Contextual & 0.963 & 0.822 & 0.853 & 0.967 & 0.832 & 0.860 & 0.004 & \cellcolor{blue!30}0.010 & 0.007 \\
cot (gpt-4) & Lexical & 0.425 & 0.348 & 0.819 & 0.569 & 0.499 & 0.859 & 0.144 & \cellcolor{blue!5}0.151 & 0.040 \\
cot (gpt-4) & Scalar & 0.392 & 0.329 & 0.839 & 0.661 & 0.563 & 0.853 & 0.269 & \cellcolor{pink!90}0.234 & 0.014 \\
\midrule
posthoc (gpt-4) & gpt-3.5 & 0.592 & 0.490 & 0.828 & 0.796 & 0.689 & 0.865 & 0.204 & \cellcolor{pink!15}0.199 & 0.037 \\
posthoc (gpt-4) & gpt-4 & 0.745 & 0.622 & 0.835 & 0.910 & 0.807 & 0.886 & 0.165 & \cellcolor{blue!15}0.185 & 0.051 \\
posthoc (gpt-4) & Llama-3.3 & 0.697 & 0.588 & 0.843 & 0.888 & 0.782 & 0.880 & 0.191 & \cellcolor{blue!5}0.194 & 0.037 \\
posthoc (gpt-4) & PresupFlip & 0.668 & 0.559 & 0.837 & 0.878 & 0.804 & 0.889 & 0.210 & \cellcolor{pink!60}0.245 & 0.052 \\
posthoc (gpt-4) & Contextual & 0.955 & 0.801 & 0.839 & 0.970 & 0.819 & 0.844 & 0.015 & \cellcolor{blue!30}0.018 & 0.005 \\
posthoc (gpt-4) & Lexical & 0.344 & 0.306 & 0.860 & 0.528 & 0.561 & 0.886 & 0.184 & \cellcolor{pink!90}0.255 & 0.026 \\
posthoc (gpt-4) & Scalar & 0.650 & 0.553 & 0.850 & 0.832 & 0.761 & 0.878 & 0.182 & \cellcolor{pink!35}0.208 & 0.028 \\
\midrule
cot (Llama-3.3) & gpt-3.5 & 0.626 & 0.528 & 0.843 & 0.813 & 0.706 & 0.869 & 0.187 & \cellcolor{pink!35}0.178 & 0.026 \\
cot (Llama-3.3) & gpt-4 & 0.796 & 0.682 & 0.857 & 0.935 & 0.821 & 0.878 & 0.139 & \cellcolor{blue!15}0.139 & 0.021 \\
cot (Llama-3.3) & Llama-3.3 & 0.751 & 0.643 & 0.855 & 0.921 & 0.799 & 0.868 & 0.179 & \cellcolor{blue!5}0.156 & 0.013 \\
cot (Llama-3.3) & PresupFlip & 0.652 & 0.564 & 0.865 & 0.896 & 0.755 & 0.843 & 0.244 & \cellcolor{pink!90}0.191 & -0.022 \\
cot (Llama-3.3) & Contextual & 0.959 & 0.449 & 0.468 & 0.968 & 0.455 & 0.470 & 0.009 & \cellcolor{blue!30}0.006 & 0.002 \\
cot (Llama-3.3) & Lexical & 0.445 & 0.342 & 0.768 & 0.627 & 0.522 & 0.811 & 0.182 & \cellcolor{pink!60}0.180 & 0.043 \\
cot (Llama-3.3) & Scalar & 0.652 & 0.532 & 0.816 & 0.832 & 0.696 & 0.843 & 0.180 & \cellcolor{pink!15}0.164 & 0.027 \\
\midrule
posthoc (Llama-3.3) & gpt-3.5 & 0.590 & 0.480 & 0.813 & 0.816 & 0.706 & 0.865 & 0.226 & \cellcolor{pink!15}0.226 & 0.052 \\
posthoc (Llama-3.3) & gpt-4 & 0.744 & 0.623 & 0.837 & 0.929 & 0.815 & 0.877 & 0.185 & \cellcolor{blue!15}0.192 & 0.040 \\
posthoc (Llama-3.3) & Llama-3.3 & 0.711 & 0.588 & 0.826 & 0.939 & 0.822 & 0.893 & 0.228 & \cellcolor{pink!90}0.234 & 0.067 \\
posthoc (Llama-3.3) & PresupFlip & 0.666 & 0.577 & 0.867 & 0.886 & 0.799 & 0.848 & 0.220 & \cellcolor{blue!5}0.222 & -0.019 \\
posthoc (Llama-3.3) & Contextual & 0.955 & 0.296 & 0.310 & 0.959 & 0.331 & 0.345 & 0.004 & \cellcolor{blue!30}0.035 & 0.035 \\
posthoc (Llama-3.3) & Lexical & 0.436 & 0.316 & 0.829 & 0.631 & 0.548 & 0.844 & 0.195 & \cellcolor{pink!60}0.232 & 0.015 \\
posthoc (Llama-3.3) & Scalar & 0.661 & 0.565 & 0.855 & 0.881 & 0.794 & 0.866 & 0.220 & \cellcolor{pink!35}0.229 & 0.011 \\
\bottomrule
\end{tabular}
}
\caption{Per-condition results for counterfactual simulatability \textit{(judge: \texttt{gpt-4.1})}. Improvements are Phase~2 minus Phase~1.}
\label{tab:cf-sim-gpt41}
\end{table*}

\section{Automatic Evaluation for Counterfactuals}

\begin{table}[h]
\centering
\resizebox{\columnwidth}{!}{%
\begin{tabular}{lllrr}
\toprule
\textbf{cfq\_type}  & \textbf{cfq\_type} & \textbf{TS} & \textbf{LFR\_gold} & \textbf{LFR\_orig} \\
\midrule
model\_generated & gpt-3.5          & 0.777 & 0.487 & 0.497 \\
model\_generated & gpt-4                  & 0.713 & 0.535 & 0.525 \\
model\_generated & Llama-3.3 & 0.634 & 0.503 & 0.517 \\
prag\_cf         & presupp\_flip   & 1.480 & \textbf{0.629} & \textbf{0.534} \\
prag\_cf         & contextual             & 1.255 & 0.350 & 0.349 \\
prag\_cf         & lexical                & 0.203 & 0.450 & 0.401 \\
prag\_cf         & scalar                 & \textbf{0.184} & 0.437 & 0.397 \\
\bottomrule
\end{tabular}%
}
\caption{Label Flip Rate and Textual Similarity scores for model-generated and pragmatics-based counterfactual questions.}
\label{tab:textual-similarity-cfa}
\end{table}

In addition to simulation-based evaluation, we also consider automated metrics that are commonly used to characterize counterfactual data. These metrics quantify how a perturbation $Q'$ related to its original question $Q$, independently of any explanation. 

\paragraph{Label Flip Rate} The label flip rate measures how often a perturbed example changes the model’s predicted label relative to the original instance \citep{Ge2021CounterfactualEF,nguyen-etal-2024-llms,Bhattacharjee2023TowardsLC,wang-etal-2025-fitcf}.
Intuitively, counterfactuals that preserve the surface form of $Q$ but flip the label are considered more challenging and informative, because they probe decision boundaries rather than trivial paraphrases.

We compute the proportion of $Q'$ for which the model’s answer differs from its answer to $Q$, and later relate this flip rate to the observed gains from showing explanations in our simulation test.

\paragraph{Textual Similarity} Textual similarity captures how close $Q'$ remains to $Q$ in form and content \citep{Madaan2020GenerateYC}. High similarity is often taken to indicate a “minimal” counterfactual, where small edits lead to different model behavior, whereas low similarity suggests that the perturbation may be drifting too far from the original instance.

Following prior work on counterfactual evaluation, we quantify textual similarity using a normalized word-level Levenshtein distance \cite{ross-etal-2021-explaining,treviso-etal-2023-crest,wang-etal-2025-fitcf}. 


Table~\ref{tab:textual-similarity-cfa} reports the Label Flip Rate and textual similarity results across counterfactual question types. However, both metrics only approximate counterfactual quality and ignore whether humans can actually leverage explanations to anticipate model behavior, which is what our simulation-based evaluation is designed to capture.



\section{Full results for in-context bias and flip-back behavior}\label{sec:results-bias-flipback}

In this appendix, we report the full results on in-context bias and flip-back behavior for explanations generated by GPT-3.5 and Llama-3.3-70B in Table~\ref{tab:bias-gpt3.5-llama}. The patterns mirror those observed for GPT-4: in-context bias generally decreases from Phase~1 to Phase~2, with especially pronounced reductions when conditioning on instances where the original answer is incorrect.

\begin{table*}[h]
\centering
\resizebox{0.85\textwidth}{!}{%
\begin{tabular}{llrrrr}
\toprule
\multirow{2}{*}{\textbf{expl\_type}} & \multirow{2}{*}{\textbf{cfq\_type}} 
& \multicolumn{2}{c}{\textbf{gpt-3.5}} 
& \multicolumn{2}{c}{\textbf{Llama-3.3}} \\
\cmidrule(lr){3-4} \cmidrule(lr){5-6}
& & \textbf{$\Delta$bias\_orig} & \textbf{$\Delta$bias\_orig\_wrong} 
  & \textbf{$\Delta$bias\_orig} & \textbf{$\Delta$bias\_orig\_wrong} \\
\midrule
\multicolumn{6}{c}{\cellcolor[HTML]{EFEFEF}\textit{\textbf{Weaker Evaluator}}} \\
cot     & gpt-3.5       & -0.072 & -0.093 & -0.044 & -0.077 \\
cot     & gpt-4         & -0.087 & -0.139 & -0.025 & -0.013 \\
cot     & Llama-3.3     & -0.087 & -0.078 & \cellcolor{blue!15}0.003  & -0.086 \\
cot     & presupp\_flip & -0.047 & -0.060 & -0.162 & -0.015 \\
cot     & contextual    & -0.023 & -0.045 & -0.015 & -0.073 \\
cot     & lexical       & -0.079 & -0.145 & -0.055 & -0.063 \\ 
cot     & scalar        & -0.029 & -0.054 & -0.093 & -0.108 \\
\midrule
posthoc & gpt-3.5       & -0.048 & -0.091 & -0.110 & -0.292 \\
posthoc & gpt-4         & -0.015 & -0.106 & -0.080 & -0.283 \\
posthoc & Llama-3.3     & -0.043 & -0.143 & -0.063 & -0.321 \\
posthoc & presupp\_flip & -0.164 & -0.013 & -0.197 & -0.011 \\
posthoc & contextual    & -0.022 & -0.038 & -0.018 & -0.062 \\
posthoc & lexical       & -0.065 & -0.119 & -0.065 & -0.083 \\
posthoc & scalar        & -0.067 & -0.083 & -0.055 & -0.069 \\
\multicolumn{6}{c}{\cellcolor[HTML]{EFEFEF}\textit{\textbf{Stronger Evaluator}}} \\
cot     & gpt-3.5       & -0.090 & -0.241 & -0.060 & -0.035 \\
cot     & gpt-4         & -0.057 & -0.201 & -0.037 & -0.041 \\
cot     & Llama-3.3     & -0.087 & -0.078 & -0.059  & -0.039 \\
cot     & presupp\_flip & -0.073 & -0.014 & -0.052  & \cellcolor{blue!15}0.011 \\
cot     & contextual    & -0.017 & -0.084 & -0.016  & \cellcolor{blue!15}0.008 \\
cot     & lexical       & -0.061 & -0.073 & -0.061  & -0.033 \\
cot     & scalar        & -0.049 & -0.010 & -0.071  & -0.053 \\
\midrule
posthoc & gpt-3.5       & -0.041 & -0.010 & -0.092 & \cellcolor{blue!15}0.028 \\
posthoc & gpt-4         & -0.035 & -0.002 & -0.071  & \cellcolor{blue!15}0.039 \\
posthoc & Llama-3.3     & -0.059 & -0.010 & -0.095  & \cellcolor{blue!15}0.040 \\
posthoc & presupp\_flip & -0.085 &  \cellcolor{blue!15}0.008 & -0.067  & \cellcolor{blue!15}0.017 \\
posthoc & contextual    & -0.036 &  \cellcolor{blue!15}0.024 & -0.045  & \cellcolor{blue!15}0.033 \\
posthoc & lexical       & -0.065 &  \cellcolor{blue!15}0.008 & -0.083  & -0.008 \\
posthoc & scalar        & -0.043 & -0.014 & -0.060  & -0.034 \\
\bottomrule
\end{tabular}
}
\caption{Change in in-context bias from Phase~1 to Phase~2 across counterfactual question sets, reported separately for gpt-3.5 and Llama-3.3-70B. Negative values indicate reduced bias (more flip-back). Positive numbers are marked in blue.}
\label{tab:bias-gpt3.5-llama}
\end{table*}

\section{Human--LLM Judge Agreement on the Overlapping Subset}\label{sec:appendix-human-llm-agreement}

To assess how well LLM-as-a-judge approximates human judgments, we compute raw yes/no agreement between human predictions and the LLM judge on the overlapping subset of items. In Table~\ref{tab:human-llm-agreement-by-type} we report agreement separately for Phase~1 (without explanations) and Phase~2 (with explanations). In addition, we provide label distributions for both raters and a breakdown by counterfactual type. These statistics complement the main text by documenting where human and LLM judge decisions align or diverge across transformation types. The full results are shown in 

\begin{table}[h]
\centering
\small
\begin{tabular}{lcc}
\toprule
\textbf{cfq\_type} & \textbf{Agreement (P1)} & \textbf{Agreement (P2)} \\
\midrule
\multicolumn{3}{c}{\cellcolor[HTML]{EFEFEF}\textit{\textbf{Weaker Evaluator}}} \\
contextualization      & 0.776 & 0.820 \\
gpt\_3.5               & 0.531 & 0.540 \\
gpt\_4                 & 0.469 & 0.440 \\
lexical\_substitution  & 0.735 & 0.740 \\
llama                  & 0.612 & 0.560 \\
pre\_flip              & 0.347 & 0.300 \\
scalar\_adjustment     & 0.735 & 0.740 \\
\multicolumn{3}{c}{\cellcolor[HTML]{EFEFEF}\textit{\textbf{Stronger Evaluator}}} \\
contextualization      & 0.660 & 0.480 \\
gpt\_3.5               & 0.560 & 0.600 \\
gpt\_4                 & 0.440 & 0.540 \\
lexical\_substitution  & 0.600 & 0.480 \\
llama                  & 0.640 & 0.740 \\
pre\_flip              & 0.340 & 0.600 \\
scalar\_adjustment     & 0.720 & 0.520 \\
\bottomrule
\end{tabular}
\caption{Raw yes/no agreement between humans and the LLM-as-a-judge by counterfactual question type on the overlapping subset.}
\label{tab:human-llm-agreement-by-type}
\end{table}

\end{document}